%% file: main.tex
\documentclass[10pt,twocolumn,letterpaper]{article}

\usepackage[pagenumbers]{cvpr}      

\usepackage{multirow}
\usepackage[table,xcdraw]{xcolor}
\usepackage[normalem]{ulem}
\useunder{\uline}{\ul}{}
\usepackage{booktabs}
\usepackage{multirow}
\usepackage{colortbl}
\usepackage{graphicx}
\usepackage[normalem]{ulem}
\usepackage{xcolor}
\usepackage{enumitem} 
\usepackage[accsupp]{axessibility}
\usepackage[dvipsnames]{xcolor}
\usepackage[margin=1in]{geometry}
\usepackage{tcolorbox} 
\usepackage{xcolor}
\tcbuselibrary{skins} 

\tcbset{
  enhanced,
  boxrule=0pt,
  arc=3pt,
  boxsep=0.5pt,
  left=1pt,right=1pt,top=0.5pt,bottom=0.5pt
}

\newtcbox{\perceptionbox}{
  on line,
  colback=RoyalBlue!12, colframe=RoyalBlue!50!black,
  fontupper=\ttfamily\fontsize{8.5}{10}\selectfont, 
  left=1pt,right=1pt,top=0.6pt,bottom=0.6pt, boxsep=0.6pt, arc=3pt, boxrule=0pt
}
\newtcbox{\reasoningbox}{
  on line,
  colback=orange!15, colframe=orange!60!black,
  fontupper=\ttfamily\fontsize{8.5}{10}\selectfont, 
  left=1pt,right=1pt,top=0.6pt,bottom=0.6pt, boxsep=0.6pt, arc=3pt, boxrule=0pt
}
\newcommand\blfootnote[1]{%
  \begingroup
  \renewcommand\thefootnote{}\footnote{#1}%
  \addtocounter{footnote}{-1}%
  \endgroup
}
\definecolor{cvprblue}{rgb}{0.21,0.49,0.74}
\usepackage[pagebackref,breaklinks,colorlinks,allcolors=cvprblue]{hyperref}

\def\paperID{42754} 
\def\confName{CVPR}
\def\confYear{2026}

\title{\textsc{VideoP2R}: Video Understanding from Perception to Reasoning}

\author{Yifan Jiang$^{1*}$, Yueying Wang$^2$, Rui Zhao$^{2\dagger}$, Toufiq Parag$^{3*}$, Zhimin Chen$^2$\\
Zhenyu Liao$^2$, Jayakrishnan Unnikrishnan$^2$\\
$^1$USC, $^2$Amazon, $^3$Keystone AI\\
{\tt\small $^1$yifjia@isi.edu,$^2$\{yueyingw,zhaori,zhiminch,zyliao,jayunn\}@amazon.com} \\ {\tt\small $^3$toufiq.parag@gmail.com}}

\begin{document}
\maketitle
\input{sec/0_abstract}

\input{sec/1_intro}

\input{sec/2_related}
\input{sec/3_data}

\input{sec/4_PA_GRPO}

\input{sec/5_experiment}

\input{sec/6_result}

\input{sec/7_conclusion}

\newpage
{
    \small
    \bibliographystyle{ieeenat_fullname}
    \bibliography{main}
}

\clearpage
\setcounter{page}{1}
\maketitlesupplementary
\input{sup/0_detailed_analysis}
\input{sup/1_experiment_setup}

\input{sup/2_ablation_study_on_judge_model}

\input{sup/3_observation_examination}

\input{sup/4_rl_training_trend}
\input{sup/5_alignment_check}

\input{sup/6_more_cases}
\input{sup/7_dataset_ablation}

\end{document}

%% file: sec/0_abstract.tex
\begin{abstract}
Reinforcement fine-tuning~(RFT), a two-stage framework consisting of supervised fine-tuning~(SFT) and reinforcement learning~(RL) has shown promising results on improving reasoning ability of large language models~(LLMs). Yet extending RFT to large video language models~(LVLMs) remains challenging. We propose \textbf{\textsc{VideoP2R}}, a novel process-aware video RFT framework that enhances video reasoning by modeling perception and reasoning as distinct processes. In the SFT stage, we develop a three-step pipeline to generate \textbf{\textsc{VideoP2R}-CoT-162K}, a high-quality, process-aware chain-of-thought~(CoT) dataset for perception and reasoning. In the RL stage, we introduce a novel process-aware group relative policy optimization~(PA-GRPO) algorithm that supplies separate rewards for perception and reasoning. Extensive experiments show that \textsc{VideoP2R} achieves state-of-the-art~(SotA) performance on six out of seven video reasoning and understanding benchmarks. Ablation studies further confirm the effectiveness of our process-aware modeling and PA-GRPO and demonstrate that model's perception output is information-sufficient for downstream reasoning. Our project page is available at \url{https://videop2r.github.io/videop2r/}.\blfootnote{$^*$Work done while at Amazon. \quad $^\dagger$Corresponding author.}
\end{abstract}

%% file: sec/1_intro.tex
\section{Introduction}
\label{sec:intro}


Understanding visual information in video has long been a core challenge in computer vision~\cite{liu2024tempcompass,fu2025video,spelke2007core,tang2025video}. Reinforcement fine-tuning~(RFT)~\cite{luong2024reft}, a two-stage framework of supervised fine-tuning~(SFT) followed by reinforcement learning~(RL), has recently emerged as a powerful approach for enhancing the reasoning capabilities of large language models (LLMs)~\cite{bai2025qwen2,achiam2023gpt,jaech2024openai}. Within the RFT framework, the SFT stage leverages Chain-of-Thought~(CoT)~\cite{wei2022chain} annotations to warm up models with explicit guidance, while the RL stage further improves reasoning quality through reward-driven optimization. Motivated by the success of RFT, recent studies~\cite{wang2025time,feng2025video,li2025videochat,chen2025versavid} have adapted RFT to video reasoning by designing diverse CoT generation pipelines and developing RL variants, with particular attention to Group Relative Policy Optimization~(GRPO)~\cite{guo2025deepseek} for its simple rule-based rewards without a critic model. These adaptations have demonstrated clear advantages over conventional SFT across multiple video benchmarks~\cite{feng2025video,wang2025videorft}.

\begin{figure*}[t]
  \centering
   \includegraphics[width=1\linewidth]{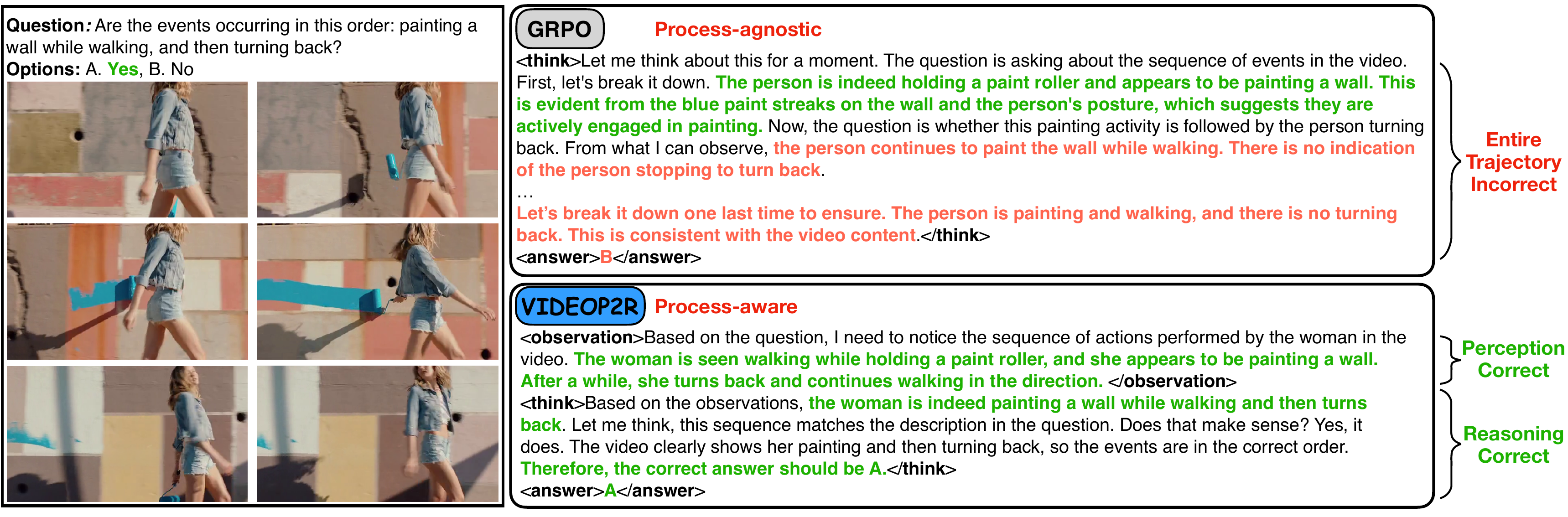}
    \caption{Comparison between GRPO-based video RFT framework~(process-agnostic) and \textsc{VideoP2R}~(process-aware).}
   \label{fig:main_figure}
\end{figure*}

However, exploration of adapting RFT from text to video remains preliminary, partly due to simply treating video as another alternative modality to text \cite{tong2024eyes}. This simplicity assumption ignores the decomposability of visual reasoning, which consists of two different processes \cite{selvaraju2020squinting,yuan2021perception}. The first one is \emph{perception}, which extracts salient information from the visual input, and \emph{reasoning}, which organizes the visual evidence and draws inferences. Each process can introduce error in distinct ways~(e.g., missed visual evidence, faulty inference), and can impact video understanding~\cite{jiang2024marvel,ahrabian2024curious}. Yet existing video RFT frameworks are process-agnostic: they collapse perception and reasoning into a single procedure and assign a single final reward to the whole trajectory, which blurs credit assignment. As \cref{fig:main_figure}~(top right) shows, a perception error~(e.g., turning back) induces the reasoning error, but without process awareness, the entire trajectory is evaluated as incorrect. 
Therefore, assigning a single reward to the entire process can prevent the model from effectively correcting mistakes that emerge in different processes.

The challenge to extend RFT to explicitly account for different processes in visual reasoning is two-fold. (i) \emph{Lack of process-aware CoT data}: existing CoT annotations conflate perception and reasoning rather than explicitly disentangling the two; and (ii) \emph{Coarse rewards}: training typically collapses feedback into a single final reward for the entire reasoning process~\cite{wang2025time,feng2025video,wang2025videorft}, hindering credit assignment across processes.
To address these challenges, we propose \textsc{\textbf{VideoP2R}}, a novel process-aware video RFT framework that models perception and reasoning as distinct processes to enhance video reasoning. Same as conventional RFT, our framework consists of two stages of training. \textbf{In the SFT stage}, we construct a three-step CoT generation pipeline that produces high-quality perception and reasoning traces. Given visual question-answer~(VQA) samples, our pipeline first generates visual perception and reasoning traces in order; the perceptions are then fed to a reasoning-capable LLM~\cite{besta2025reasoning} to verify whether they contain sufficient visual evidence to reach the correct answer. Running this pipeline on 260K VQA pairs~\cite{feng2025video} yields \textbf{\textsc{VideoP2R}-CoT-162K} after filtering low-quality samples. We use this dataset in SFT to warm up the model, encouraging process separation during inference and providing a strong initialization for subsequent RL. \textbf{In the RL stage}, we propose \textbf{PA-GRPO}, a process-aware variant of GRPO. Unlike GRPO, which assigns a single reward to the entire trajectory, PA-GRPO supplies two separate rewards specific for perception and reasoning and assigns them to the corresponding output segments: 1) an LLM-judged~\cite{gu2024survey} perception reward, which evaluates whether the model’s perception captures the necessary information from the video, and 2) a rule-based reasoning reward verifying the correctness of the final answer. As illustrated in \cref{fig:main_figure}, after two-stage training, \textsc{VideoP2R} enables LVLMs to conduct process-aware inference with calibrated perception and reasoning.

We conduct comprehensive experiments on seven widely used video understanding and reasoning benchmarks~\cite{liu2024tempcompass,zhao2025mmvu,li2024mvbench,qi2025vcr,hu2025video,fu2025video,yang2025thinking}, comparing \textsc{VideoP2R} with representative process-agnostic video RFT baselines~(e.g., Video-R1~\cite{feng2025video} and VideoRFT~\cite{wang2025videorft}). Results show that \textsc{VideoP2R} achieves SotA on six out of seven benchmarks, with robust gains of 1.9\%–9.1\% average accuracy over base models across benchmarks. Ablation studies further validate the effectiveness of the process-aware modeling and the PA-GRPO. In addition, we provide fine-grained analysis of \textsc{VideoP2R}’s perception and PA-GRPO’s improvements over GRPO to support future process-aware research in the video domain. Our key contributions are: (i) A novel process-aware video RFT framework, \textbf{\textsc{VideoP2R}}, that models perception and reasoning separately to enhance video reasoning. (ii) A process-aware RL algorithm based on GRPO, \textbf{PA-GRPO}, that provides separate rewards for perception and reasoning, improving credit assignment in RL. (iii) An automatic three-step CoT generation pipeline that produces perception and reasoning annotations, yielding \textbf{\textsc{VideoP2R}-CoT-162K} for warm start in SFT. (iv) Comprehensive evaluation confirms \textsc{VideoP2R}'s SotA performance, with ablations verifying the effectiveness of process-aware modeling and PA-GRPO. 


\begin{figure*}[t]
  \centering
   \includegraphics[width=1\linewidth]{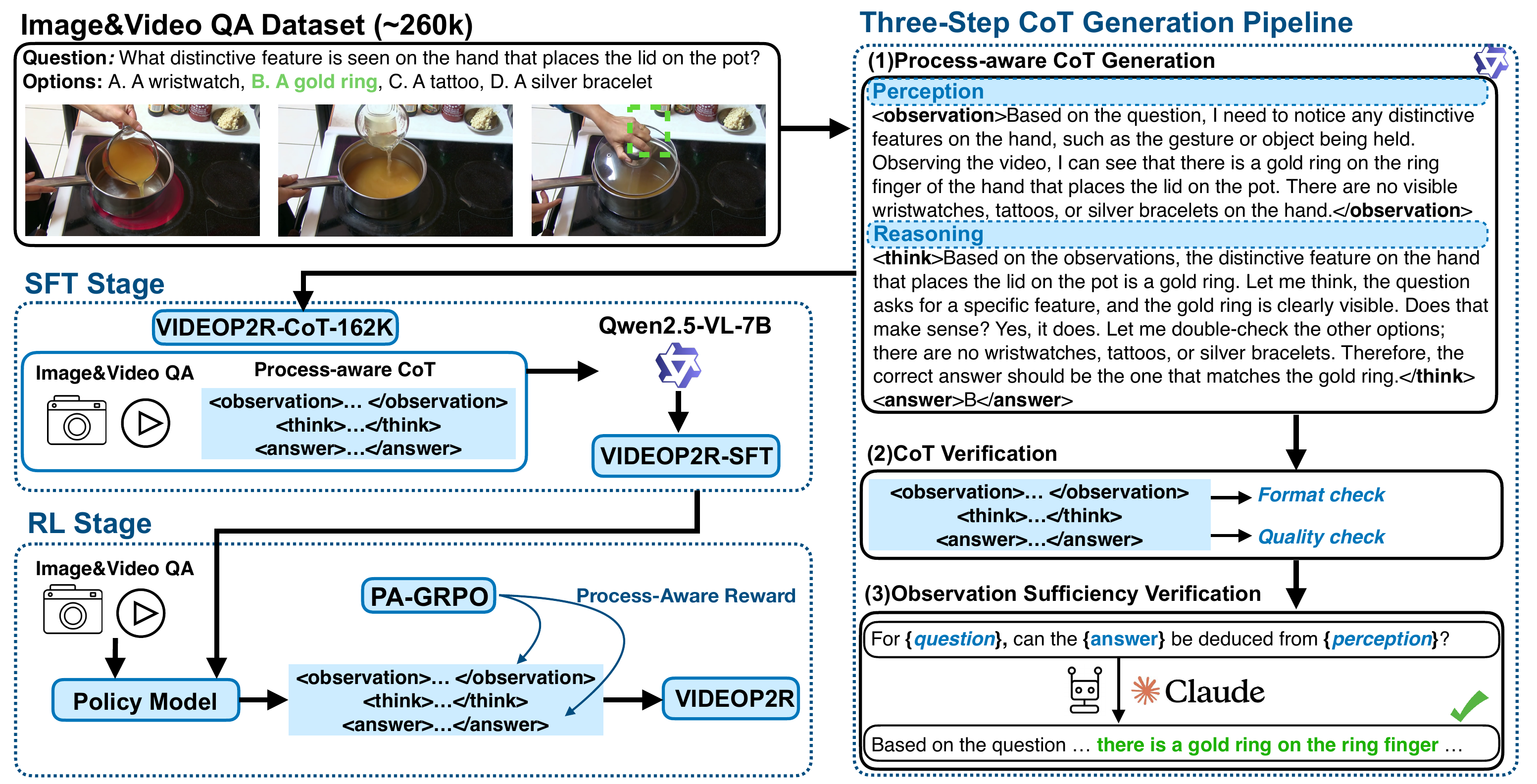}
    \caption{Illustration of overall \textsc{VideoP2R} RFT framework~(left) and the three-step CoT generation pipeline~(right).}
 \label{fig:overall_framework_annotation}
\end{figure*}

%% file: sec/2_related.tex
\section{Related work}

\label{sec:related}
\textbf{Video Perception and Reasoning in LVLMs.} Video reasoning poses coupled spatial–temporal challenges beyond text/image only settings~\cite{liu2024tempcompass,fu2025video,spelke2007core}, demanding coordinated perception and reasoning~\cite{selvaraju2020squinting,yuan2021perception}. Early approaches handle this by introducing modular perception preprocessors~(e.g., frame captioners~\cite{wang2022language,wang2024videoagent,zhang2023simple} or spatio-temporal scene graphs~\cite{songmodularized,xiao2022video,ligraph2025}) to aid downstream reasoning. However, because these preprocessors are typically frozen, they bottleneck the pipeline and cannot be improved to curb error propagation or information loss~\cite{tong2024eyes}. Recent methods organize perception–reasoning into pre-defined stages, either by attributes (frames/objects/actions)~\cite{wang2025videochat,min2024morevqa,wang2024stair,han2025videoespresso,fei2024video,qiu2025step} or steps (planning/grounding)~\cite{liu2025videomind,shi2025enhancing}. However, rigid designs often limit generalization across scenarios~\cite{wang2025videorft}, and mixing perception and reasoning at each stage makes it vulnerable to perception errors~\cite{jiang2024marvel,ahrabian2024curious}. In contrast, \textsc{VideoP2R} separates perception and reasoning, aligning with the two-process modeling of video reasoning~\cite{selvaraju2020squinting} for better generalization, while preserving an end-to-end training pipeline for continual refinement.



\textbf{Reinforcement Learning for LVLMs.} Building on GRPO, recent work adapts RL/RFT to video understanding and reasoning~\cite{park2025deepvideo,wang2025time,feng2025video,li2025videochat,zhang2025tinyllava,wang2025videorft,chen2025versavid,meng2025videocap}. Time-R1 uses timestamp-aware and template rewards~\cite{wang2025time}; Video-R1 and STAR-R1 reward sensitivity to correct temporal order~\cite{feng2025video,li2025star}; Videochat-R1 and VersaVid-R1 adopt task-specific rewards~\cite{li2025videochat,chen2025versavid}; VideoRFT adds a stage-aware semantic reward~\cite{wang2025videorft}. However, most prior efforts model video reasoning as a single trajectory and apply a single final reward over the entire output sequence, with no process-aware distinction between perception and reasoning. This also holds for recent image-domain studies on perception–reasoning separation~\cite{xia2025visionary,li2025self,yu2025perception}. \textsc{VideoP2R} instead uses PA-GRPO to assign separate perception and reasoning rewards to their token segments, providing more clear signals and more precise error attribution.

%% file: sec/3_data.tex
\section{\textsc{VideoP2R} RFT Framework}

In this section, we introduce the overall design of the \textsc{VideoP2R} RFT framework~(\cref{fig:overall_framework_annotation} left), which follows the standard RFT setup~\cite{luong2024reft} with a specific focus on modeling video reasoning into perception and reasoning: \textbf{(1)~SFT stage},  we use a three-step CoT generation pipeline to construct a process-aware CoT dataset, \textbf{\textsc{VideoP2R}-CoT-162K}. We train the base model on this dataset to enhance its perception and reasoning capabilities while warming up the model for the RL stage. \textbf{(2)~RL stage}, we propose a process-aware reinforcement learning scheme, \textbf{PA-GRPO}, which refines the model’s reasoning by providing separate rewards for perception and reasoning, enabling the model to move beyond supervised learning boundaries~\cite{ouyang2022training}.
\subsection{Process-aware CoT Dataset}
To address the challenge of lacking process-aware CoT dataset, we develop a strategy to curate CoT data at scale, which is then used for fine-tuning LVLM in SFT.  Particularly, we first standardize  a process-aware CoT template that explicitly disentangles perception from reasoning in different segments as follows: \makebox[\linewidth][c]{
  \perceptionbox{\texttt{<observation>\dots</observation>}}
}
\makebox[\linewidth][c]{
  \reasoningbox{\texttt{<think>\dots</think><answer>\dots</answer>}}
}
$\langle \texttt{observation}\rangle$ segment represents the perception process, where the model extracts relevant visual evidence based on the question. $\langle \texttt{think}\rangle$ and $\langle \texttt{answer}\rangle$ segment captures the reasoning process, where the model reasons~($\langle \texttt{think}\rangle$) over the extracted visual evidence and states the final answer~($\langle \texttt{answer}\rangle$).  All generated CoTs follow this template, and the model is trained to adhere to it at inference.

\subsubsection{Three-Step CoT Generation Pipeline}
Building upon the proposed template, we design a three-step pipeline~(\cref{fig:overall_framework_annotation} right) to generate process-aware CoT data containing both perception and reasoning traces for a diverse set of VQA tasks.
The overall workflow comprises three major steps~(Details in the Supplementary).
\textbf{1) Process-aware CoT Generation.} For each VQA sample, we use Qwen2.5-VL-72B-Instruct~\cite{bai2025qwen2} to generate an initial CoT trace for both perception and reasoning in corresponding segments.
\textbf{2) CoT Verification.} 
To ensure consistency and correctness, we evaluate the final answer of each generated response with task-specific metrics~(e.g., exact word match and word error rate for generation tasks), discarding samples that yield low-quality answers or deviate from the expected CoT template. 
\textbf{3) Observation Sufficiency Verification.} 
We further filter generated data using a cross-modal validation strategy to validate the $\langle\texttt{observation}\rangle$ segment in isolation from raw visual inputs~\cite{wang2025videorft}. Specifically, for each sample, we provide only the $\langle \texttt{observation}\rangle$ segment, along with the corresponding question and answer, to Claude 3.7 Sonnet~\cite{claude3team2023anthropic}, which assesses whether the visual evidence in $\langle \texttt{observation}\rangle$ are adequate to support the final reasoning process.

\subsubsection{Data Construction and Statistics}
To ensure reproducibility and fair comparison, we apply this pipeline to a public and representative image- and video-based QA dataset~\cite{feng2025video}, which includes multiple question types such as multiple-choice, numerical QA, Optical Character Recognition~(OCR), free-form QA, and regression.
Applying our pipeline on 260K VQA data~\cite{feng2025video} produces 162K high-quality process-aware CoT data with perception and reasoning traces, termed \textbf{\textsc{VideoP2R}-CoT-162K}. 
We provide detailed analyses of \textsc{VideoP2R}-CoT-162K~(e.g.,  embedding visualization and word-frequency statistics) in the Supplementary, which show that our annotations inherently separate perception from reasoning.

%% file: sec/4_PA_GRPO.tex
\subsection{Process-Aware Reinforcement Learning}
\begin{figure}[t]
  \centering
   \includegraphics[width=1\linewidth]{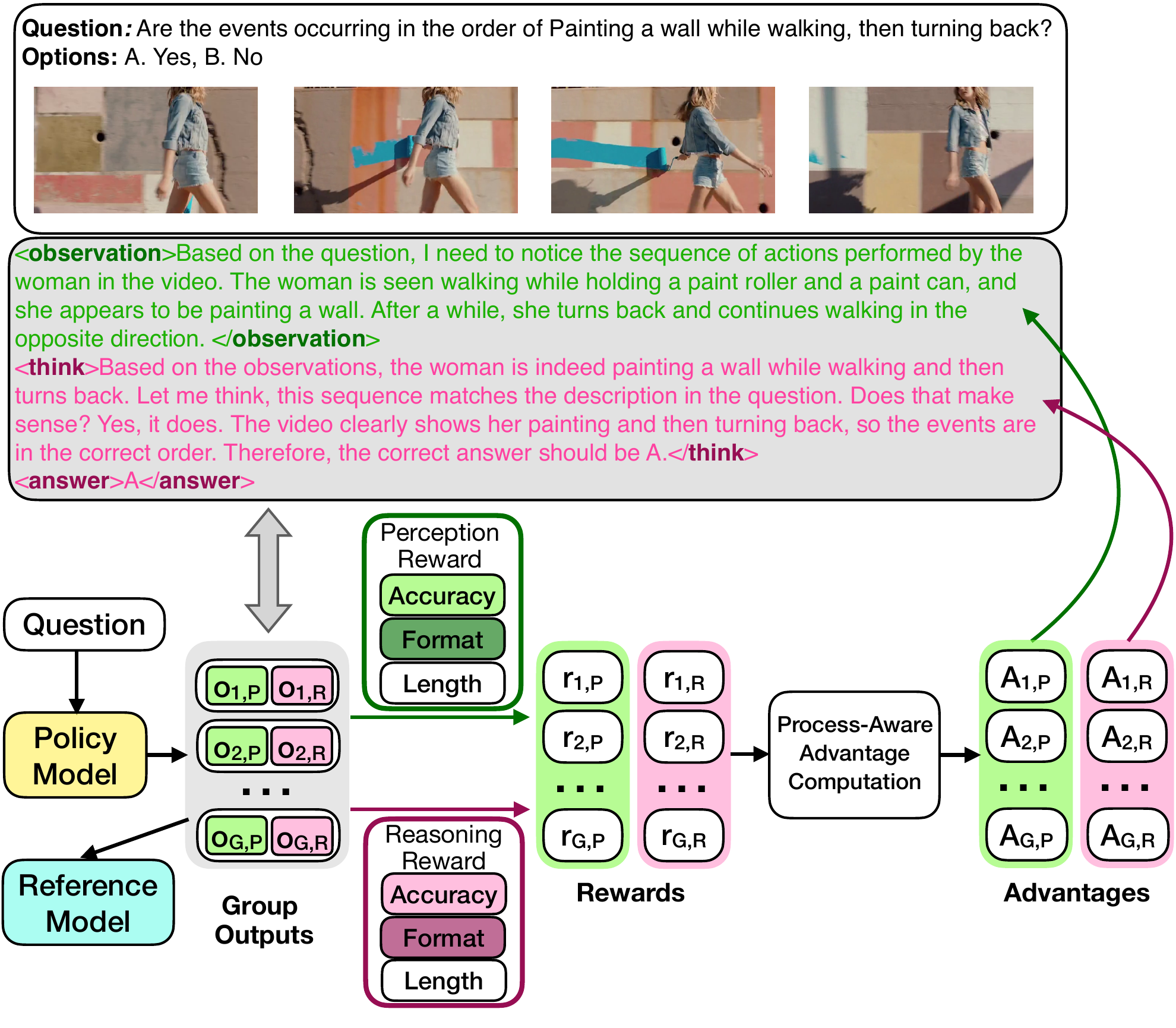}

   \caption{The illustration of the PA-GRPO algorithm.}
   \label{fig:PA-GRPO}
\end{figure}

After the SFT stage, we further refine the model through reinforcement learning. Building upon GRPO~\cite{guo2025deepseek}, we introduce a process-aware variant, \textbf{PA-GRPO}~(\cref{fig:PA-GRPO}), which provides separate reward signals for perception and reasoning processes to encourage more structured and efficient policy optimization~\cite{lightman2023let}.
This section first revisits the standard GRPO framework and then presents our process-aware extension designed to align reward signals with perception–reasoning separation.
\subsubsection{Group Relative Policy Optimization~(GRPO)}

GRPO~\cite{guo2025deepseek} extends Proximal Policy Optimization~\cite{schulman2017proximal} by removing the dependency on a learned critic model and directly comparing responses within groups. 
Given a question $q$, the policy model $\pi_{\theta}$ samples $G$ candidate responses $o = \{o_1, o_2, \dots, o_G\}$ as a group, each assigned a rule-based reward $r_i$. 
Rewards are then normalized within the group to yield the relative advantage:
\begin{equation}
A_i = 
\frac{r_i - \mathrm{mean}(\{r_j\}_{j=1}^G)}
{\mathrm{std}(\{r_j\}_{j=1}^G)}.
\label{eq:advantage}
\end{equation}
where $A_i$ represents the relative advantage of all tokens in the $i$-th response within the group. With the relative advantage computed, the GRPO overall optimization objective is formulated as:
\begin{align} 
\mathcal{J}_{\text{GRPO}}(\theta) 
&= 
\mathbb{E}_{q, \{o_i\}}\!\left[ 
\frac{1}{G}\sum_{i=1}^{G} 
\min\!\left( \rho_i A_i,\; 
\right.\right. \nonumber 
\left.\left. 
\mathrm{clip}\!\big( 
\rho_i,\, 
1\!-\!\epsilon,\right.\right. \nonumber\\[-2pt] &\hspace{0.2cm} 
\left.\left. 
1\!+\!\epsilon 
\big)A_i \right) 
-\beta\,\mathbb{D}_{\mathrm{KL}}\!\big(\pi_{\theta}\,\|\,\pi_{\text{ref}}\big) 
\right]. 
\label{eq:grpo} 
\end{align}
where $\rho_i = \frac{\pi_{\theta}(o_i\mid q)}{\pi_{\theta_{\text{old}}}(o_i\mid q)}$ 
is the likelihood ratio between updated policy $\pi_{\theta}$ and old policy $\pi_{\theta_{\text{old}}}$, and $\pi_{\text{ref}}$ is a fixed reference model~(e.g., a frozen copy of policy model) after SFT, providing KL regularization weighted $\beta$.
This formulation constrains large policy deviations while promoting high-reward samples, ensuring stable optimization during reinforcement learning.

\subsubsection{Process-Aware Group Relative Policy Optimization~(PA-GRPO)}

While GRPO performs well on textual reasoning, its single scalar reward provides a limited training signal for the two-process based visual reasoning~\cite{selvaraju2020squinting,yuan2021perception}.
To address this, we propose \textbf{PA-GRPO}, which introduces separate \textbf{perception} and \textbf{reasoning} rewards for tokens in each process, enabling fine-grained credit assignment during reinforcement learning, as illustrated in~\cref{fig:PA-GRPO}.
Formally, for each question $q$, the sampled response from the policy model $\pi_{\theta}$ is represented as
$o = \{(o_{1,\text{P}}, o_{1,\text{R}}), (o_{2,\text{P}}, o_{2,\text{R}}), \dots, (o_{G,\text{P}}, o_{G,\text{R}})\}$,
where $o_{i,\text{P}}$ denotes perception process tokens within the $\langle \texttt{observation} \rangle$ segment, and $o_{i,\text{R}}$ denotes reasoning process tokens within the $\langle \texttt{think} \rangle$ and $\langle \texttt{answer} \rangle$ segments. For tokens within each process, \textbf{PA-GRPO} supplies separate accuracy rewards to provide reliable supervision. We further introduce length and format rewards to encourage well-structured, concise outputs, following prior RFT frameworks~\cite{guo2025deepseek,feng2025video,wang2025videorft,li2025videochat}.
We demonstrate the accuracy reward design, and our configurations for the format and length rewards below.\\
\noindent\textbf{Perception Accuracy Reward~($R_\text{acc,P}$).} The perception accuracy reward evaluates whether the model correctly perceives visual information from video input. We adopt an LLM-as-Judge evaluation~\cite{gu2024survey} in a text-only setting, a procedure shown to be reliable in various scenarios~\cite{chiang2024chatbot,dubois2024length}. Concretely, we provide only the 
$\langle \texttt{observation}\rangle$ segment, along with the corresponding question and answer, to Claude 3.7 Sonnet~\cite{claude3team2023anthropic}, which judges whether the 
$\langle \texttt{observation}\rangle$ segment is sufficient to supcont the correct answer.
Formally,
\begin{equation}
R_{\text{acc},\text{P}} = 1~\text{(judged sufficient); } 0~\text{otherwise.}
\end{equation}

\noindent\textbf{Reasoning Accuracy Reward~($R_\text{acc,R}$).} The reasoning accuracy reward evaluates whether the model produces accurate reasoning outcomes. We apply task-specific evaluation metrics to accommodate different question types, including exact word match for categorical tasks, ROUGE-based similarity for open-ended generation, and error-based scores for numerical or regression problems. Formally,
\begin{equation}
R_{\text{acc,R}} = \text{Acc}_{t}(o_{i,\text{R}}, y_{\text{true}}),
\end{equation}
where $\text{Acc}_{t}(\cdot)\in[0,1]$ denotes the task-specific accuracy metric for task type $t$.


\noindent\textbf{Format Reward~($R_{\text{form}}$) and Length Reward~($R_{\text{len}}$).} To ensure clear reward assignment, the format and the length reward in PA-GRPO are also provided separately for each process~($o_{i,\text{P}}$ and $o_{i,\text{R}}$).
We use the format reward~($R_{\text{form}}$) to encourage adherence to the process-aware inference template. Specifically, perception process tokens~($o_{i,\text{P}}$) must appear within the $\langle \texttt{observation}\rangle$, the reasoning process tokens~($o_{i,\text{R}}$) must present the reasoning trace under $\langle \texttt{think}\rangle$, and the final answer must be provided within $\langle \texttt{answer}\rangle$. We verify compliance using regular expression matching and assign a binary reward~(\{0,1\}) accordingly. The length reward~($R_{\text{len}}$) is included to favor concise yet informative responses while avoiding overthinking. The reward is assigned only if both accuracy and format rewards are non-zero and the model’s response in each segment~($o_{i,\text{P}}$ and $o_{i,\text{R}}$) falls within the target length~([$l_{\text{min}}$,\ $l_{\text{max}}$]). In line with prior RFT work~\cite{wang2025videorft,feng2025video}, we fix \(R_{\text{len}}=0.2\), $l_{\text{min}} = 128$ and $l_{\text{max}} = 320$ for $o_{i,\text{P}}$, $l_{\text{min}} = 320$ and $l_{\text{max}} = 512$ for $o_{i,\text{R}}$.

\subsubsection{Process-Aware Reward Assignment}
The overall reward of perception tokens~($o_{i,\text{P}}$) and reasoning tokens ($o_{i,\text{R}}$) are defined as 
\begin{equation}
R_{i,k} = R_{\text{acc},k} + R_{\text{form},k} + R_{\text{len},k},
\quad k \in \{\text{P}, \text{R}\}.
\label{eq:reward_pr}
\end{equation}
Unlike Eq.~(\ref{eq:advantage}), which normalizes all rewards within a single group, we split rewards into separate groups~(perception process vs. reasoning process) and normalize each to get process-aware advantage, since their scales and distributions are not directly comparable~\cite{zeng2025reinforcing}:
\begin{equation}
A_{i,k} = 
\frac{R_{i,k} - \mathrm{mean}(\{R_{j,k}\}_{j=1}^G)}
{\mathrm{std}(\{R_{j,k}\}_{j=1}^G)}, 
\quad k \in \{\text{P}, \text{R}\}.
\label{eq:advantage_pr}
\end{equation}

We assign each process-aware advantage only to its corresponding tokens~(e.g., the perception advantage~$A_{i,\text{P}}$ is applied to~$o_{i,\text{P}}$). The overall optimization objective of PA-GRPO is then formulated as:
\begin{align} 
\mathcal{J}_{\text{PA-GRPO}}(\theta) 
&= 
\mathbb{E}_{q, \{o_i\}}\!\left[ 
\frac{1}{G}\sum_{i=1}^{G} 
\sum_{k \in \{\text{P},\text{R}\}} 
\min\!\left( 
\rho_{i,k} A_{i,k},\; 
\right.\right. \nonumber\\[-2pt] 
&\hspace{-1.7cm} 
\left.\left. 
\mathrm{clip}\!\big( 
\rho_{i,k},\, 
1\!-\!\epsilon,\,
1\!+\!\epsilon 
\big)A_{i,k} 
\right) 
-\beta\,\mathbb{D}_{\mathrm{KL}}\!\big(\pi_{\theta}\,\|\,\pi_{\text{ref}}\big) 
\right], 
\label{eq:pa-grpo} 
\end{align}
where $\rho_{i,k} = \dfrac{\pi_{\theta}(o_{i,k}\mid q)}{\pi_{\theta_{\text{old}}}(o_{i,k}\mid q)}$ 
denotes the likelihood ratio for perception ($k{=}\text{P}$) or reasoning ($k{=}\text{R}$), 
and $A_{i,k}$ is the process-aware advantage.

%% file: sec/5_experiment.tex
\begin{table*}[!t]
  \caption{Performance comparison on video reasoning and understanding benchmarks. Best/second-best result of each column is in \textbf{bold}/\underline{underline}. Missing entries indicate unreported results (all numbers unit in \%).}
  \vspace{-2mm} %
  \centering
  \small
  \setlength{\tabcolsep}{6pt}
  \renewcommand{\arraystretch}{0.98}
  \begin{tabular}{l|rrrr|rrr|c}
    \toprule
    \multirow{2}{*}{Model} & \multicolumn{4}{c|}{Video Reasoning} & \multicolumn{3}{c|}{Video Understanding} & \multirow{2}{*}{Avg} \\
    \cmidrule(lr){2-5}\cmidrule(lr){6-8}
    & VSI. & VideoMMMU & MMVU & VCR. & MV. & TempCom. & VideoMME & \\
    \midrule
    \rowcolor[HTML]{F8F9FB}
    \multicolumn{9}{l}{\textbf{Open-Source 7B Models}}\\[-1pt]
    LLaVA-OneVision-7B  & 32.4 & 33.8 & 49.2 & -- & 56.7 & --  & 58.2 & -- \\
    LongVA-7B           & 29.2 & 23.9 & --   & -- & --   & 56.9 & 52.6 & -- \\
    Video-UTR-7B        & --   & --   & --   & -- & 58.8 & 59.7 & 52.6 & -- \\
    VideoLLaMA2-7B      & --   & --   & 44.8 & -- & 54.6 & --   & 47.9 & -- \\
    Qwen2.5-VL-7B       & 30.1 & 48.1 & 60.0 & 44.3 & 59.0 & 72.6 & 56.6 & 52.9 \\
    \midrule
    \rowcolor[HTML]{F8F9FB}
    \multicolumn{9}{l}{\textbf{RFT on Qwen2.5-VL-7B}}\\[-1pt]
    Video\text{-}R1        & \underline{35.8} & 52.3 & 63.8 & 49.0 & 63.9 & 73.2 & 59.3 & 56.8 \\
    VideoChat\text{-}R1    & 33.9 & \underline{54.0} & 63.0 & 49.0 & \underline{67.9} & 72.5 & 57.7 & 56.9 \\
    Time\text{-}R1         & 29.0 & 51.0 & 62.9 & 49.6 & 63.1 & 73.7 & 59.3 & 55.5 \\
    VersaVid\text{-}R1     & 33.7 & 51.9 & 64.3 & \underline{49.8} & 62.9 & \underline{74.0} & 58.8 & 56.5 \\
    VideoRFT               & \textbf{36.8} & 51.1 & \textbf{68.5} & 49.6 & 62.1 & 73.7 & \underline{59.8} & \underline{57.4} \\
    \midrule
    \rowcolor[HTML]{E7FAFE}
    \textsc{VideoP2R} (Ours)      & \textbf{36.8} & \textbf{55.0} & \underline{65.4} & \textbf{51.0} & \textbf{68.1} & \textbf{74.5} & \textbf{60.0} & \textbf{58.7} \\
    \bottomrule
  \end{tabular}
  \label{tab:main_results}
  \vspace{-2mm} %
\end{table*}

\section{Experiment Setup}
\textbf{Two-stage Training.}
We adopt Qwen2.5-VL-7B-Instruct~\cite{bai2025qwen2} as the base LVLM in our training pipeline. Following the same training setups used in prior video-RFT studies~\cite{feng2025video,wang2025videorft}, we perform one epoch of SFT on \textsc{VideoP2R}-CoT-162K, followed by 1K RL updates over original 260K visual QA data using PA-GRPO. The model obtained after the SFT stage is referred to as \textbf{\textsc{VideoP2R}-SFT}, and the final model after the RL stage is denoted as \textbf{\textsc{VideoP2R}}.\\
\noindent\textbf{Benchmarks.}
Following prior works~\cite{fu2025video,wang2025videorft}, we evaluate our approach on seven benchmarks, including four video reasoning datasets (VSI-Bench~\cite{yang2025thinking}, VideoMMMU~\cite{hu2025video}, MMVU~\cite{zhao2025mmvu}, and VCR-Bench~\cite{qi2025vcr}) and three video understanding datasets (MVBench~\cite{li2024mvbench}, TempCompass~\cite{liu2024tempcompass}, and VideoMME~\cite{fu2025video}).
These benchmarks jointly cover spatial reasoning, knowledge-intensive QA, temporal logic, and general video understanding. 
We follow the official evaluation protocols of each benchmark.\\
\noindent\textbf{Baselines.}
(1)~\textbf{RFT on Qwen2.5-VL-7B.} We  compare against recent video RFT approaches built upon Qwen2.5-VL-7B using GRPO or its variants, including Video-R1~\cite{feng2025video}, Time-R1~\cite{wang2025time}, VideoRFT~\cite{wang2025videorft}, VideoChat-R1~\cite{li2025videochat}, and VersaVid-R1~\cite{chen2025versavid}~(2)~\textbf{Open-Source Models.} We further include Qwen2.5-VL-7B along with other 7B-scale models for a comprehensive evaluation:~LLaVA-OneVision~\cite{li2024llava}, LongVA~\cite{zhang2024long}, Video-UTR~\cite{yu2025unhackable}, and VideoLLaMA2~\cite{cheng2024videollama}.
We follow the prompt templates in each baseline's official publication.\\
We provide all setup details in the Supplementary.

%% file: sec/6_result.tex
\section{Results}
Our experiments focus on addressing five research questions from Section \ref{sec: main_result} to \ref{sec: qualitative_results}
: (1)~How does  \textsc{VideoP2R} perform across various video understanding benchmarks? (2)~What is the contribution of each component in \textsc{VideoP2R}?  (3)~Does the perception representations of \textsc{VideoP2R} effectively support downstream reasoning? (4)~Does the process-aware reward design of PA-GRPO improve RL efficiency and reliability? (5)~What are the success and failure mode of \textsc{VideoP2R}?

\subsection{Main Results}
\label{sec: main_result}
The main evaluation results of \textsc{VideoP2R} and other baselines  are shown in \cref{tab:main_results}. Compared with prior video RFT approaches, \textsc{VideoP2R} achieves highly competitive performance across seven benchmarks, setting \textbf{SotA results on six of them} and ranking second on the remaining one. In contrast to previous RFT methods that often bring improvement on specific datasets~(e.g., Video\text{-}R1 ranks second on VSI-Bench, while VideoRFT is SotA on MMVU but last on MVBench), \textsc{VideoP2R} delivers \textbf{consistent gains across all benchmarks}, surpassing the previous SotA by 1.3\% in average accuracy. This consistency underscores the effectiveness and generalizability of modeling visual reasoning through distinct perception and reasoning processes. Compared to the base model Qwen2.5-VL, \textsc{VideoP2R} exhibits clear and steady improvements, with accuracy gains ranging from \textbf{1.9\% to 9.1\%} across benchmarks.
More broadly, a performance gap exists between models trained with RFT and those trained only with SFT/Instruction Tuning~\cite{liu2023visual}~(i.e., open-source models), highlighting the superiority of RFT for expanding capability boundaries.  
We further analyze \textsc{VideoP2R}’s performance drop on MMVU~(\cref{sec: qualitative_results}) and attribute it to the lack of domain-specific knowledge~(e.g., chemistry) in our training data.

\begin{table*}[!t]
\caption{Ablation studies of \textsc{VideoP2R} on two-stage training, process-aware modeling and reward design (all numbers unit in \%).}
  \centering
  \small
  \setlength{\tabcolsep}{6pt}
  \renewcommand{\arraystretch}{0.98}
  \scalebox{1}[0.95]{ 
  \begin{tabular}{l|rrrr|rrr|l}
    \toprule
    \addlinespace[-1pt] 
    \multicolumn{9}{l}{\footnotesize \textit{Ablation Factor}}\\[-2pt]
    \multirow{2}{*}{\quad\quad Model Variant} & \multicolumn{4}{c|}{Video Reasoning} & \multicolumn{3}{c|}{Video Understanding} & \multirow{2}{*}{Avg.} \\
    \cmidrule(lr){2-5}\cmidrule(lr){6-8}
     & VSI. & VideoMMMU & MMVU & VCR. & MV. & TempCom. & VideoMME & \\
    \midrule 
    \addlinespace[-1pt] 
    \multicolumn{9}{l}{\footnotesize \textit{Two-stage Training}}\\[-1pt]
    \rowcolor[HTML]{E7FAFE} - \textsc{VideoP2R}~(Ours) & 36.8 & \textbf{55.0} & \textbf{65.4} & \textbf{51.0} & \textbf{68.1} & \textbf{74.5} & \textbf{60.0} & \textbf{58.7}$_{\textcolor{ForestGreen}{\scriptsize +5.8}}$ \\
    - SFT-only~(\textsc{VideoP2R}-SFT) & 35.2 & 53.7 & 61.6 & 46.9 & 62.3 & 72.4 & 57.2 & 55.6$_{\textcolor{ForestGreen}{\scriptsize +2.7}}$ \\
    - RL-only & 35.8 & 54.6 & 64.6 & 46.3 & 60.8 & 73.8 & 55.9 & 56.0$_{\textcolor{ForestGreen}{\scriptsize +3.1}}$
 \\
    \midrule
    \addlinespace[-1pt] 
    \multicolumn{9}{l}{\footnotesize \textit{Process-aware Modeling}}\\[-1pt]
        \rowcolor[HTML]{E7FAFE} - \textsc{VideoP2R}~(Ours) & 36.8 & \textbf{55.0} & \textbf{65.4} & \textbf{51.0} & \textbf{68.1} & \textbf{74.5} & \textbf{60.0} & \textbf{58.7}$_{\textcolor{ForestGreen}{\scriptsize +5.8}}$ \\
    - process-agnostic RL~(GRPO) & 37.4 & 53.6 & 62.8 & 48.3 & 63.8 & 73.3 & 55.4 & 56.4$_{\textcolor{ForestGreen}{\scriptsize +3.5}}$
 \\
    - process-aware SFT~(no RL) & 35.2 & 53.7 & 61.6 & 46.9 &
  62.3 & 72.4 & 57.2 & 55.6$_{\textcolor{ForestGreen}{\scriptsize +2.7}}$
 \\
    - process-agnostic SFT~(no RL) & 34.3 & 48.9 & 61.6 & 47.3 & 59.0 & 69.7 & 54.0 & 53.5$_{\textcolor{ForestGreen}{\scriptsize +0.6}}$
 \\
    \midrule
    \addlinespace[-1pt]  
    \multicolumn{9}{l}{\footnotesize \textit{Reward Design}}\\[-1pt]
    \rowcolor[HTML]{E7FAFE} - \textsc{VideoP2R}~(Ours) & 36.8 & \textbf{55.0} & \textbf{65.4} & \textbf{51.0} &
      \textbf{68.1} & \textbf{74.5} & \textbf{60.0} & \textbf{58.7}$_{\textcolor{ForestGreen}{\scriptsize +5.8}}$ \\
    - without $R_{\text{R}}$ & 36.0 & 51.6 & 60.3 & 46.8 & 62.1 & 72.5 & 57.9 & 55.3$_{\textcolor{ForestGreen}{\scriptsize +2.4}}$
 \\
    - without $R_{\text{P}}$ & 37.4 & 53.6 & 62.8 & 48.3 & 63.8 & 73.3 & 55.4 & 56.4$_{\textcolor{ForestGreen}{\scriptsize +3.5}}$
 \\
    - without $R_{\text{L}}$ & \textbf{40.0} & 52.7 & 63.2 & 48.4 & 65.5 & 73.9 & \textbf{60.0} & 57.7$_{\textcolor{ForestGreen}{\scriptsize +4.8}}$
 \\
    - without $separation$ & 37.1 & 53.2 & 64.9 & 48.8 & 65.0 & 73.2 & 59.7 & 57.4$_{\textcolor{ForestGreen}{\scriptsize +4.5}}$
 \\
    \midrule
    Baseline: Qwen2.5-VL-7B & 30.1 & 48.1 & 60.0 & 44.3 & 59.0 & 72.6 & 56.6 & 52.9 \\
    \bottomrule
  \end{tabular}
  }

\vspace{-1mm}
\label{tab:ablation}
\end{table*}

\subsection{Ablation Study}
\label{sec: ablation}
The success of \textsc{VideoP2R} underscores the importance of decomposing visual reasoning into distinct process stages. To further analyze the contribution of each process-aware component, we perform an ablation study on the \textbf{two-stage training} in the RFT framework, \textbf{process-aware modeling}, and \textbf{reward design}~(\cref{tab:ablation}). \\
\noindent\textbf{Two-stage Training:} SFT-only and RL-only improves the baseline by 2.7\% and 3.1\% respectively. But combining both yield a more significant 5.8\% improvement, suggesting that single-stage training is insufficient, whereas using both stages can further extend the model’s capability.\\
\noindent\textbf{Process-aware Modeling:}
We evaluate process-aware modeling against a process-agnostic counterpart in both SFT and RL. In SFT, the process-aware variant~(same as \textsc{VideoP2R}-SFT) yields an average accuracy gain of 2.1\% over the process-agnostic variant, which uses reasoning-only segments. In RL, the agnostic variant follows GRPO—assigning a single reasoning reward across all output tokens, while the process-aware variant~(same as \textsc{VideoP2R}) again leads on six benchmarks with 2.3\% on average. Under identical visual inputs, adding perception annotations in SFT and using process-aware credit assignment in RL provide clearer training signals, improving video understanding and reasoning.\\
\noindent\textbf{Reward design:}
We examine the PA-GRPO reward function by ablating each reward component and the separation reward assignment. Removing any component causes a notable drop, sometimes even below the SFT baseline~(e.g., removing the perception reward yields worse performance than SFT on VideoMME).
This indicates that all components are necessary and that their joint design in PA-GRPO enables fine-grained credit assignment in RL. When we remove separation and assign both perception and reasoning rewards to all output tokens, the results remain competitive but still lag behind PA-GRPO, indicating that separation reward assignment provides clearer and more effective training signals. Notably, removing $R_{\text{len}}$ improves performance on VSI-Bench. Further analysis~(Supplementary) indicates that its questions often require long, fine-grained descriptions to ensure sufficient perception, where the length reward becomes counter-productive. We propose to have a
dynamic length reward ~\cite{wan2025srpo} in future work.
Finally, we further ablate different judge models in the Supplementary for perception accuracy reward assignment, and find that \textsc{VideoP2R} remains robust.

\begin{figure}[!h]
  \centering
   \includegraphics[width=1\linewidth]{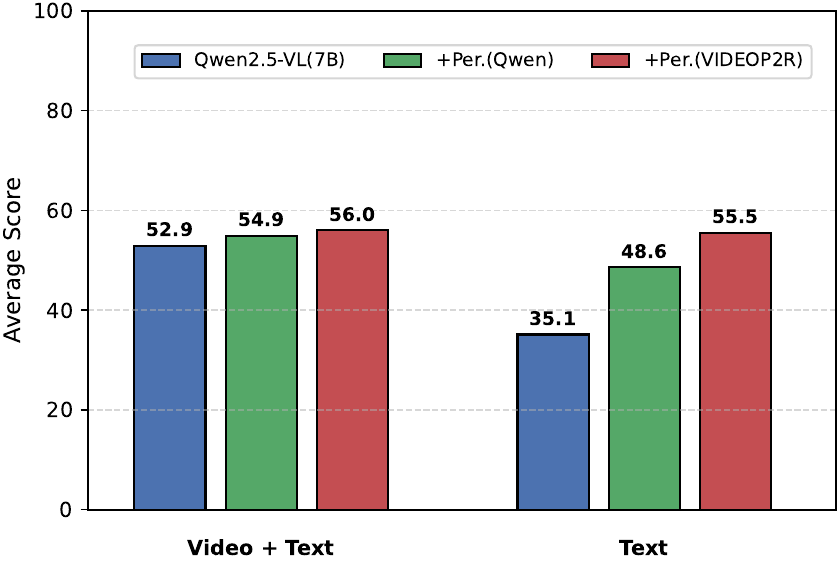}

   \caption{Effect of perception on downstream reasoning}
   \label{fig:observation}
\end{figure}

\subsection{Effectiveness of Perception Representations}
\label{sec: effect_perception}
The ablation study highlights the effectiveness of process-aware modeling in \textsc{VideoP2R}. In this section, we further examine whether the perception outputs produced by \textsc{VideoP2R} can enhance the reasoning ability of generic LVLMs~(Examination details in the Supplementary). Specifically, we compare Qwen2.5-VL-7B’s zero-shot performance on (i) text-only questions, (ii) text plus video inputs. For each scenario, we further augment question text with a perception segment generated either by \textsc{VideoP2R} or by pre-trained Qwen. \\
The results, as shown in \cref{fig:observation}, yield two key insights.
\textbf{First}, when Qwen2.5-VL-7B is provided only with text questions augmented by \textsc{VideoP2R}'s perceptions segments, its performance~(55.5\%) even surpasses that under raw video input~(52.9\%), indicating that the textual perceptions generated by \textsc{VideoP2R} capture semantically rich perceptual information that directly supports reasoning. 
\textbf{Second}, in both text and video modality, when we augment the prompt with a perception segment, using \textsc{VideoP2R}'s perceptions consistently outperforms using Qwen's own perceptions, indicating that \textsc{VideoP2R}'s perceptions provide more accurate and useful evidence for downstream reasoning.

\begin{figure*}[t]
  \centering
   \includegraphics[width=1\linewidth]{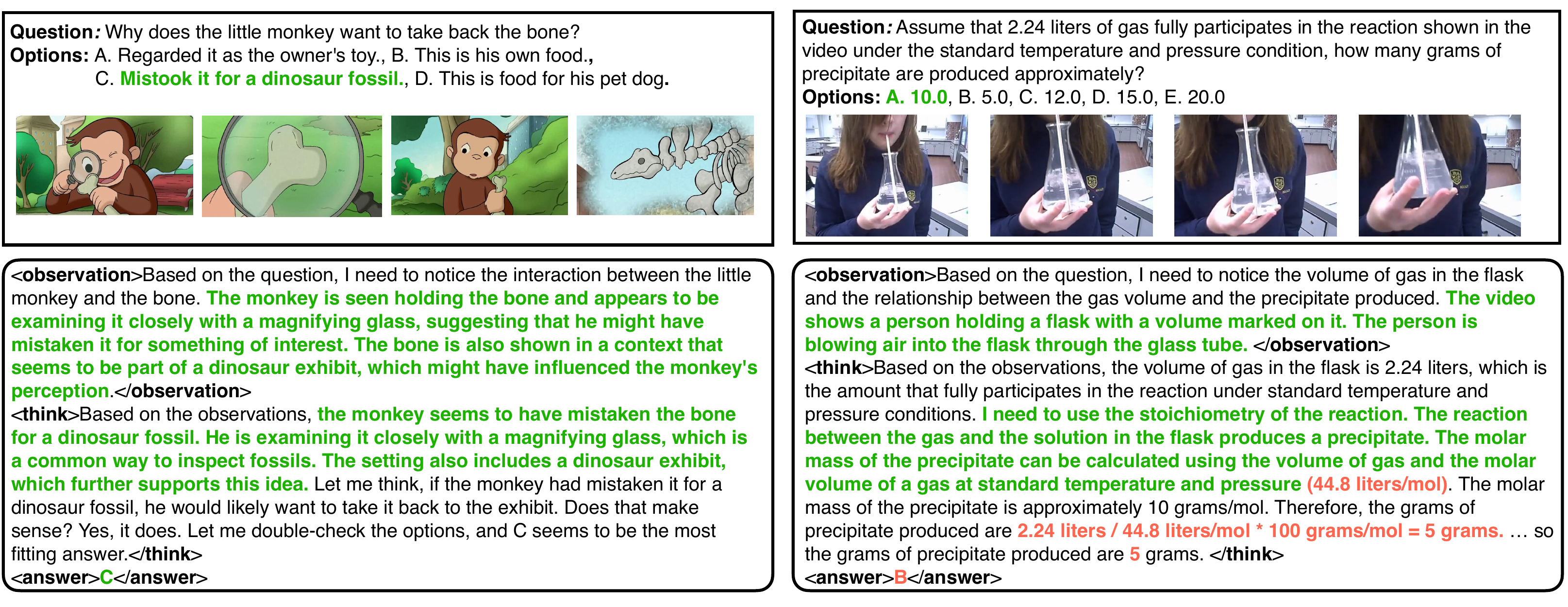}
    \caption{Success~(Left) and Failure~(Right) case of \textsc{VideoP2R}. \textcolor{ForestGreen}{Correct}  statement and \textcolor{red}{incorrect} statement are colored.}
   \label{fig:case_study}
\end{figure*}

\subsection{Advantages of PA-GRPO over GRPO}
\label{sec: advantage_of_PAGRPO}
\begin{figure}[t]
  \centering
   \includegraphics[width=1\linewidth]{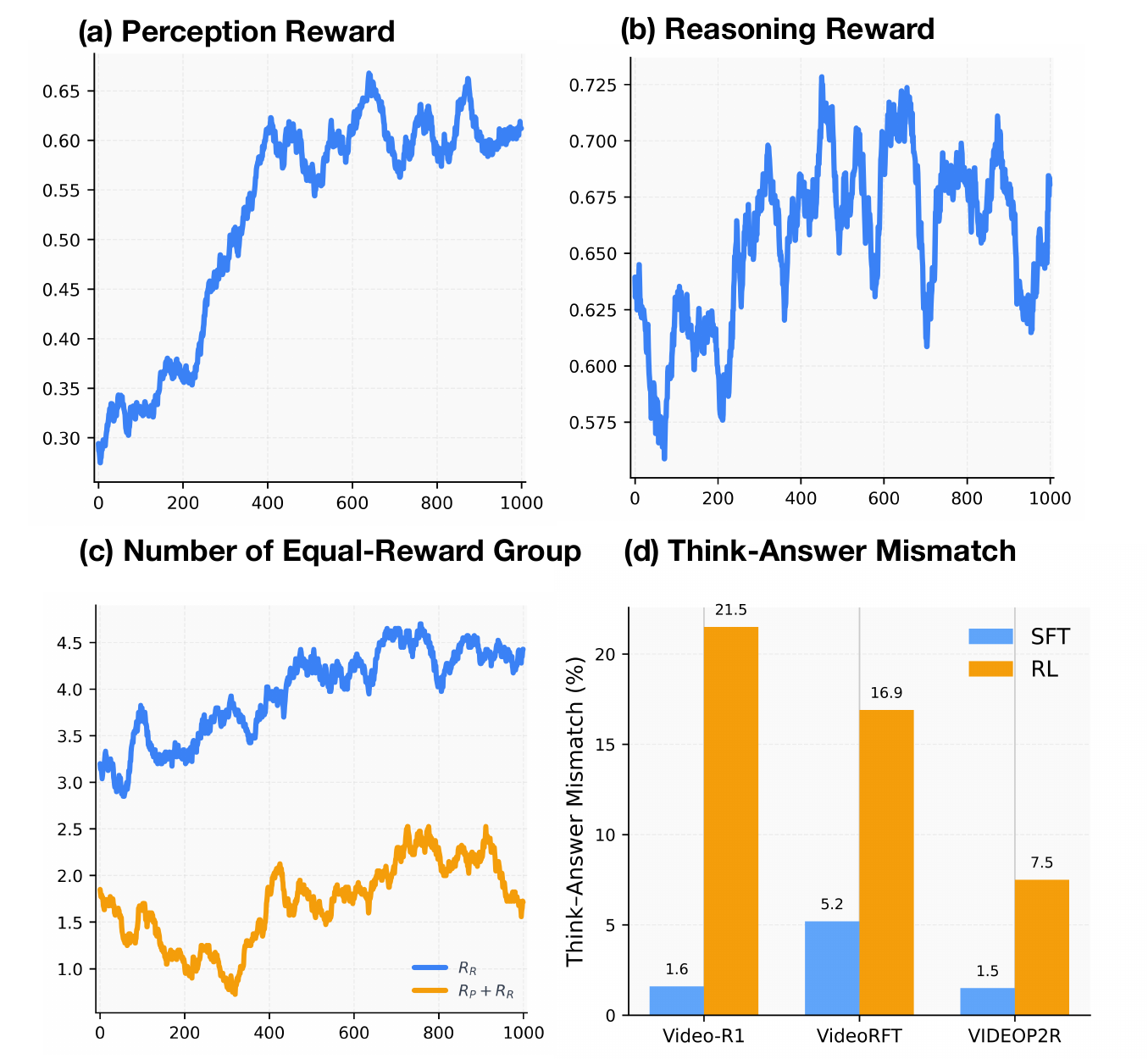}

   \caption{Training Dynamics and Think-Answer Mismatch Analysis of \textsc{VideoP2R}. Details in Section \ref{sec: advantage_of_PAGRPO}.}
   \label{fig:reward_analysis}
\end{figure}
\Cref{fig:reward_analysis}(a-b) illustrates the training dynamics of \textsc{VideoP2R}. Both perception and reasoning accuracy rewards exhibit an increasing trend, which, combined with the quantitative gains in \cref{tab:ablation}, confirms that PA-GRPO is more effective than standard GRPO in improving the model’s perception and reasoning traces. In the following, we further analyze the underlying reasons for PA-GRPO’s advantages over standard GRPO. \textbf{(1)~Training Efficiency.} In advantage-based policy gradient methods~\cite{wang2020truly,schulman2017proximal,guo2025deepseek}, including GRPO, when all sampled responses~(${o_1, o_2, \dots, o_G}$) for a given prompt receive nearly identical rewards, this leads to \textit{advantage collapse}~\cite{zhang2025edge}, where the advantages shrink toward zero, leaving little to no effective learning signal and causing updates to stagnate~\cite{yu2025dapo}. PA-GRPO mitigates this by decomposing the overall reward into perception and reasoning components, so even when reasoning rewards saturate, perception rewards can still provide non-zero gradients.
We visualize the number of samples in a batch with \textit{advantage collapse} for PA-GRPO and GRPO during the RL stage in \cref{fig:reward_analysis}(c). Compared with GRPO, PA-GRPO consistently exhibits fewer advantage collapse samples, indicating better utilization of training samples and improved training efficiency.\\
\textbf{(2) Mitigating Think–Answer Mismatch.}
Reasoning-augmented models often exhibit \textit{Think–Answer Mismatch}~\cite{shen2025mitigating}, where generated reasoning traces diverge from the actual decision process yet still produce correct answers~(e.g., “\textit{the man is on the right side, thus the answer is B. left}”).
In GRPO, such inconsistencies can lead to reward hacking~\cite{christiano2017deep,skalse2022defining}, as a single final reward can reinforce unfaithful reasoning traces that coincidentally yield correct outcomes.
To quantify this issue, we perform an alignment check~(Details in Supplementary) using Claude 3.7 Sonnet to extract answers from the $\langle \texttt{think} \rangle$ segments and compare them with the final output answer.
\Cref{fig:reward_analysis}(d) reports mismatch rates across \textsc{VideoP2R} and two single-reward trained models: while all SFT models maintain stable reasoning consistency~($\leq$5\%), both Video-R1 and VideoRFT's RL models show significant degradation~($\geq$16\%).
In contrast, \textsc{VideoP2R} shows notably lower mismatch, indicating that PA-GRPO’s process-aware rewards, which separately encourage faithful perception traces and correct final answers, effectively mitigate Think–Answer Mismatch and promote more reliable reasoning.


\subsection{Qualitative Results of \textbf{\textsc{VideoP2R}}}
\label{sec: qualitative_results}
We present one success and one failure case of \textsc{VideoP2R} in \cref{fig:case_study} to illustrate both its strengths and areas of improvement~(More examples in the Supplementary).
The left example shows an \textit{Aha Moment}~\cite{guo2025deepseek}, where \textsc{VideoP2R} performs process-aware inference by accurately describing visual cues, such as the monkey's actions and its imagination, and reasoning over them to reach the correct answer. In contrast, the right example depicts a failure case: although the model correctly identifies relevant visual details about the person and her behavior, it produces an incorrect conclusion due to missing domain-specific knowledge~(e.g., the molar volume of a gas should be 22.4).
Overall, while \textsc{VideoP2R} exhibits strong capabilities in general video understanding and reasoning, its performance can be further improved by injecting factual and domain-specific knowledge.

%% file: sec/7_conclusion.tex
\section{Conclusion}
In this work, we introduced \textsc{VideoP2R}, a process-aware RFT framework that models perception and reasoning as distinct processes for video understanding. Through a three-step CoT generation pipeline, we constructed \textsc{VideoP2R}-CoT-162K, a large-scale process-aware dataset enabling fine-grained supervision in the SFT stage. In the RL stage, we proposed PA-GRPO, a process-aware extension of GRPO that provides separate rewards for perception and reasoning to improve credit assignment. Experiments across seven benchmarks demonstrate SotA performance and strong generalization, while ablations verify the effectiveness of process-aware modeling and PA-GRPO.


%% file: sup/0_detailed_analysis.tex
We provide additional details and illustrations for our main content in the following section: 
\begin{itemize}
    \item Process-Aware CoT Generation~(\cref{sup:data_analysis}).
    \item Evaluation Setup~(\cref{sup:experiment_detilas})
    \item Ablation Study on Judge Model~(\cref{sup:abltaion_judge})
    \item Perception Effectiveness Experiment~(\cref{sup:perception examination})
   \item RL Training Trend~(\cref{sup:rl training trend}).
   \item Think-Answer Mismatch~(\cref{sup:think-answer}).
    \item More Qualitative Results~(\cref{sup:more_qualitative results}).
    \item Impact of Model Size and Dataset Composition~(\cref{sup:dataset_contribution})
\end{itemize}

\section{Details of Process-Aware CoT Generation and Data Analysis}
\label{sup:data_analysis}

\subsection{Prompt Used}
\begin{figure}[h]
  \centering
   \includegraphics[width=0.95\linewidth]{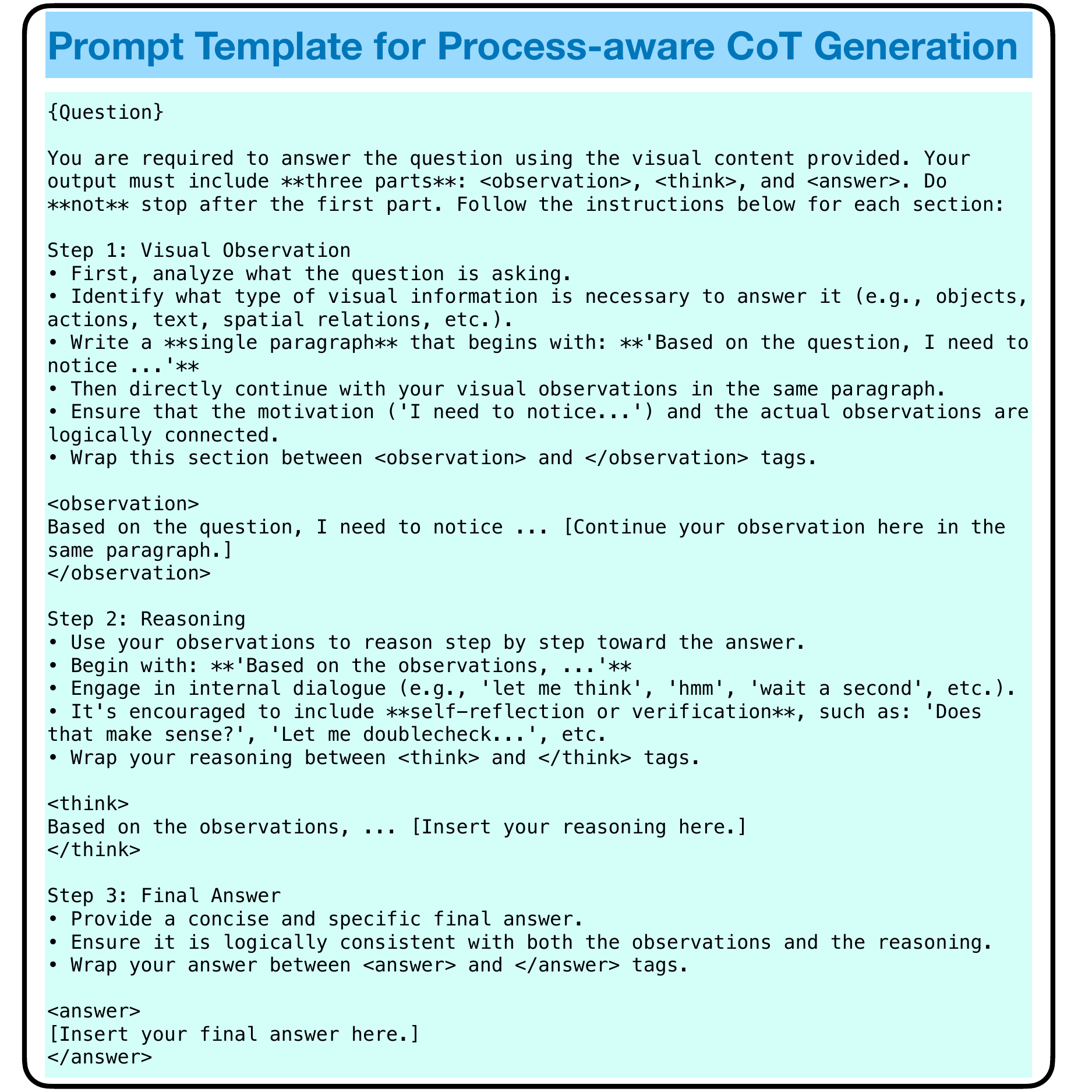}

   \caption{Prompt Template for Process-aware CoT Generation. We use the same prompt for training and inference.}
   \label{fig:prompt_for_cot}
\end{figure}

\Cref{fig:prompt_for_cot} illustrates the prompt template for process-aware CoT generation. We employ Qwen2.5-VL-72B-Instruct with a temperature of 0 for the generation.
\label{sec:prompt}

\begin{figure}[h]
  \centering
   \includegraphics[width=0.95\linewidth]{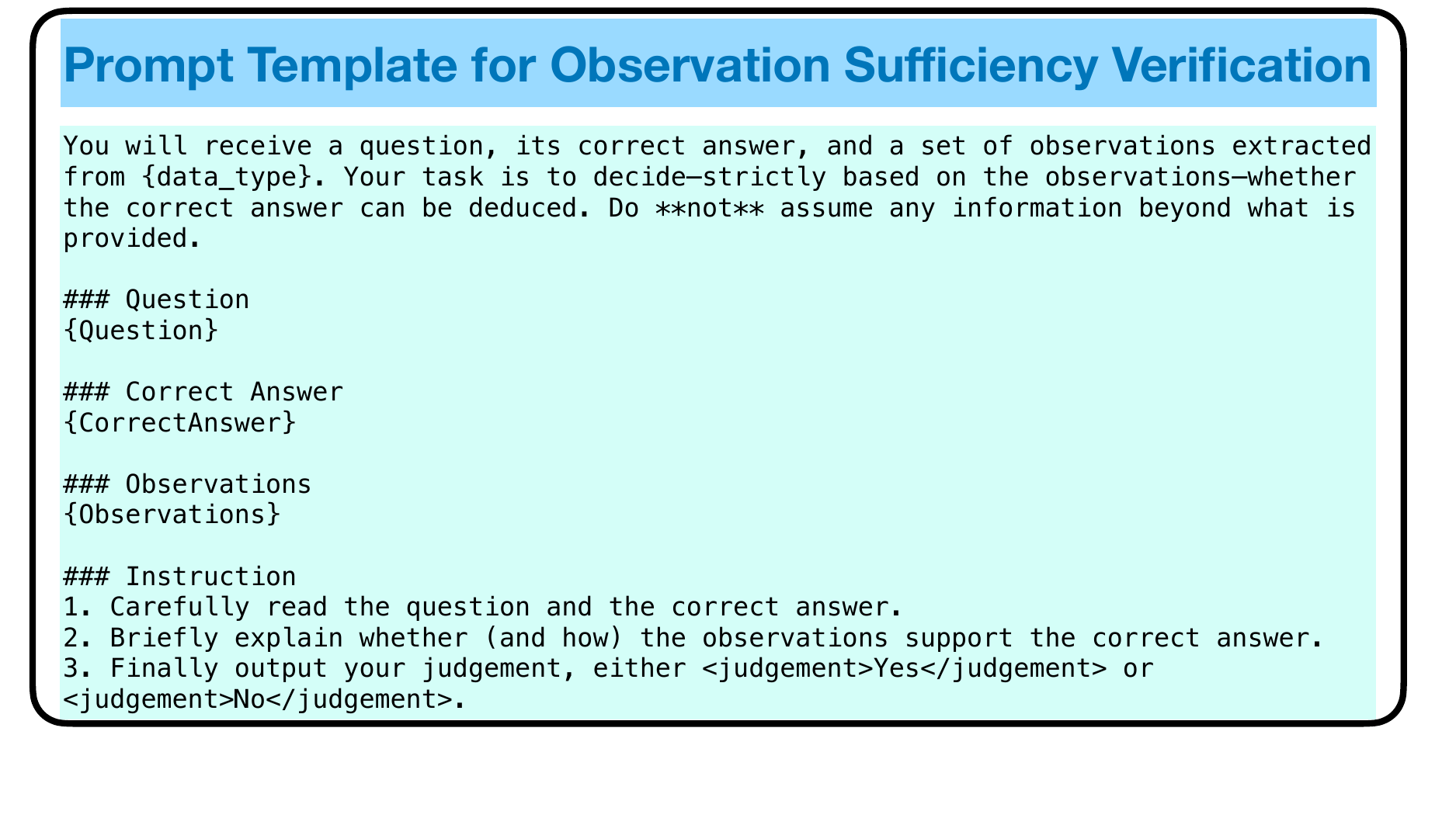}

   \caption{Prompt Template for Observation Sufficiency Verification. We use the same prompt for perception correctness judgment in RL stage.}
   \label{fig:prompt_for_obser}
\end{figure}

\Cref{fig:prompt_for_obser} illustrates the prompt template for observation sufficiency verification. We use Claude 3.7 to judge the sufficiency of the observation segment.

\subsection{Data Source and Metric for CoT Verification}

Our data source~\cite{feng2025video} encompasses five distinct question types to enhance the model’s flexibility and its generalization across diverse tasks and formats: (1) Multiple Choice, (2) Numerical QA, (3) OCR, (4) Free-form QA, and (5) Regression. Each data sample includes a question, data source, correct answer, and optional choices when applicable (e.g., for multiple-choice questions).  \\
In the subsequent CoT Verification stage, task-specific accuracy metrics are adopted to assess annotation reliability, and samples below a preset threshold of ~0.6 are filtered out. The task-specific metrics are listed as follows:
\begin{itemize}
    \item \textit{Multiple Choice}: $1$ if the predicted option matches the ground truth; $0$ otherwise.
    \item \textit{Numerical QA}: $1$ for exact match with the reference value.
    \item \textit{OCR}: reward based on Word Error Rate (WER) between prediction and reference.
    \item \textit{Free-form QA}: reward is the average of ROUGE-1, ROUGE-2, and ROUGE-L scores.
    \item \textit{Regression}: reward $= 1 -$ relative error between prediction and ground truth.
\end{itemize}
These task-specific metrics are also used for computing the reasoning accuracy reward in the RL stage.

\subsection{Data Statistic}
Adapting our generation pipeline to the data source yields \textsc{VideoP2R}-CoT-162K, consisting of 162{,}062 image and video visual QA pairs with high-quality annotations on perception and reasoning. We present the data statistics in \cref{tab:distribution analysis}. The dataset covers both image and video modalities, and spans multiple question types including multiple-choice, numerical, OCR, free-form, and regression. Multiple-choice questions constitute the majority, providing stable evaluation signals, while the inclusion of numerical, OCR, and free-form questions introduces diverse reasoning skills such as counting, reading, grounded description, and open-ended inference. This heterogeneous composition enables comprehensive assessment of process-aware perception and reasoning across modalities.

\begin{figure}[h]
  \centering
   \includegraphics[width=0.80\linewidth]{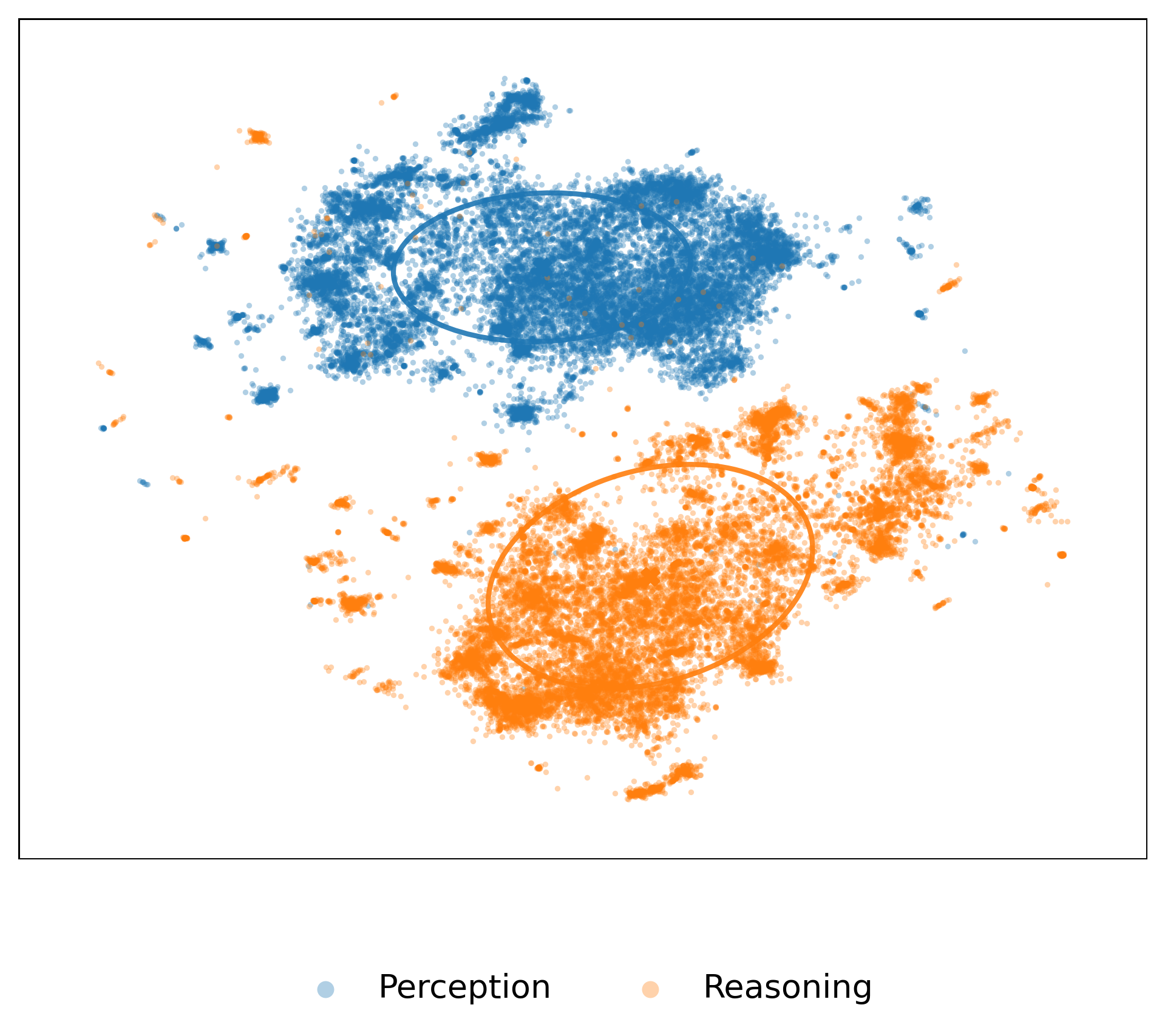}

   \caption{Embeddings visualization of \textsc{VideoP2R}-CoT-162K}
   \label{fig:embedding}
\end{figure}

To analyze our constructed data, we visualize the embedding distributions~(using UMAP~\cite{mcinnes2018umap}) of perception and reasoning annotations~(\cref{fig:embedding}). The two clusters are clearly separated, indicating that our annotated data inherently distinguishes perception and reasoning.

\begin{table*}[t]
  \caption{Distribution of question types across \textsc{VideoP2R}-CoT-162K.}
  \centering
  \small
  \setlength{\tabcolsep}{6pt}
  \renewcommand{\arraystretch}{1.05}
  \begin{tabular}{l|rrrrr|r}
    \toprule
    & \multicolumn{5}{c|}{Question Type} & \multirow{2}{*}{Sum} \\
    \cline{2-6}
    & Multiple Choice & Numerical & OCR & Free-form & Regression & \\
    \midrule
    Image & 47{,}091 & 18{,}476 & 4{,}014 & 2{,}501 & 693 & 72{,}775 \\
    Video & 86{,}910 & 1{,}371 & -- & 1{,}006 & -- & 89{,}287 \\
    \midrule
    Sum   & 134{,}001 & 19{,}847 & 4{,}014 & 3{,}507 & 693 & 162{,}062 \\
    \bottomrule
  \end{tabular}
  \label{tab:distribution analysis}
\end{table*}

\subsection{Word Count and Word Cloud Analysis}

\begin{figure}[h]
  \centering
   \includegraphics[width=0.95\linewidth]{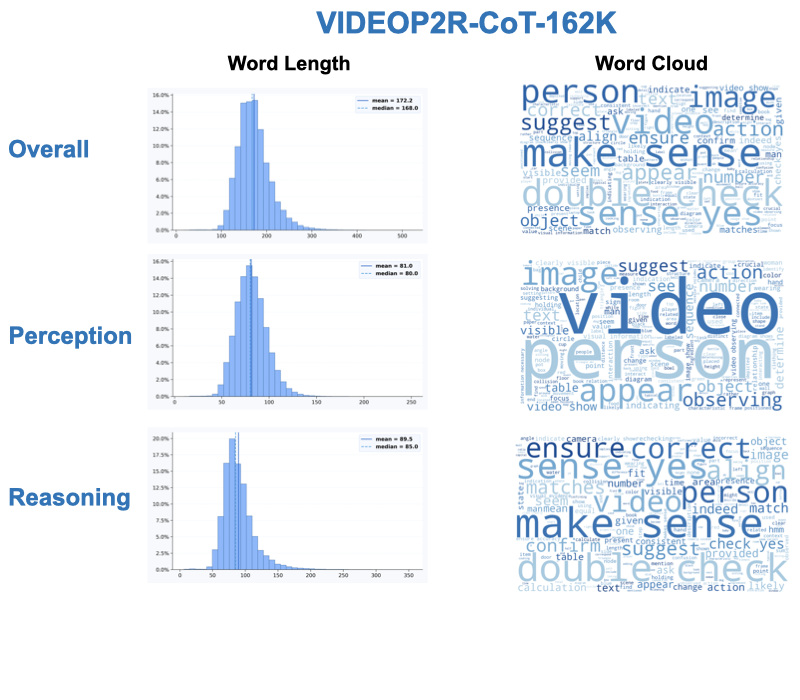}

   \caption{Word length~(Left) and Word cloud~(Right) Visualization for \textsc{VideoP2R}-CoT-162K.}
   \label{fig:detail_data}
\end{figure}

\Cref{fig:detail_data} presents the word length distribution and word cloud visualization of \textsc{VideoP2R}-CoT-162K. As shown on the left of the figure, perception and reasoning annotations exhibit a comparable number of words across the entire annotation set, suggesting a balanced contribution of both processes. The word clouds further highlight the intrinsic difference in focus between the two processes: perception annotations are dominated by video-centric terms such as “video”, “person”, and “observing”, reflecting their emphasis on factual and descriptive content; in contrast, reasoning annotations frequently contain introspective expressions such as “double check” and “make sense”, which indicate deeper reflective reasoning.

\subsection{Annotation Examples}
\begin{figure}[h]
  \centering
   \includegraphics[width=0.95\linewidth]{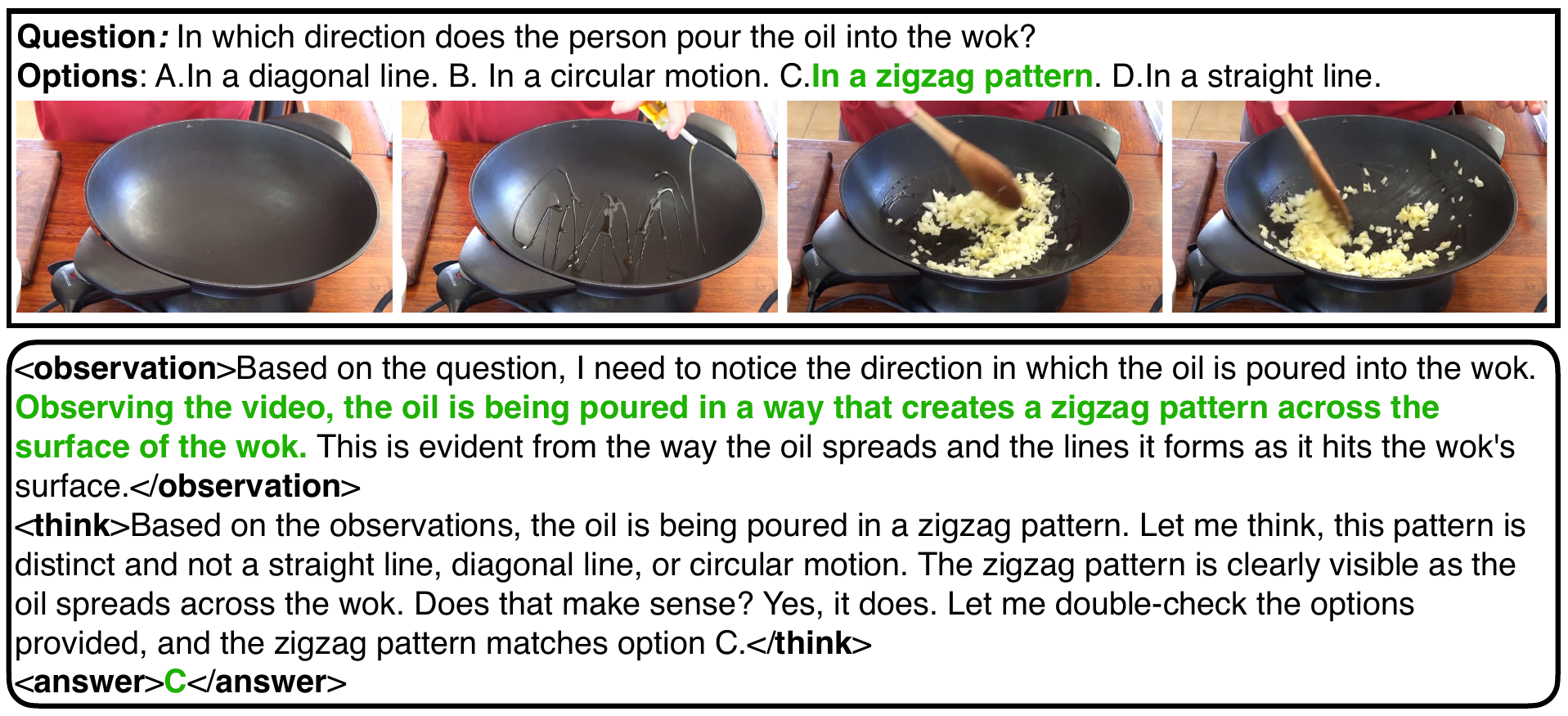}

   \caption{An Annotation Example of the Video QA Sample}
   \label{fig:annotation_sample_video}
\end{figure}

\begin{figure}[h]
  \centering
   \includegraphics[width=0.95\linewidth]{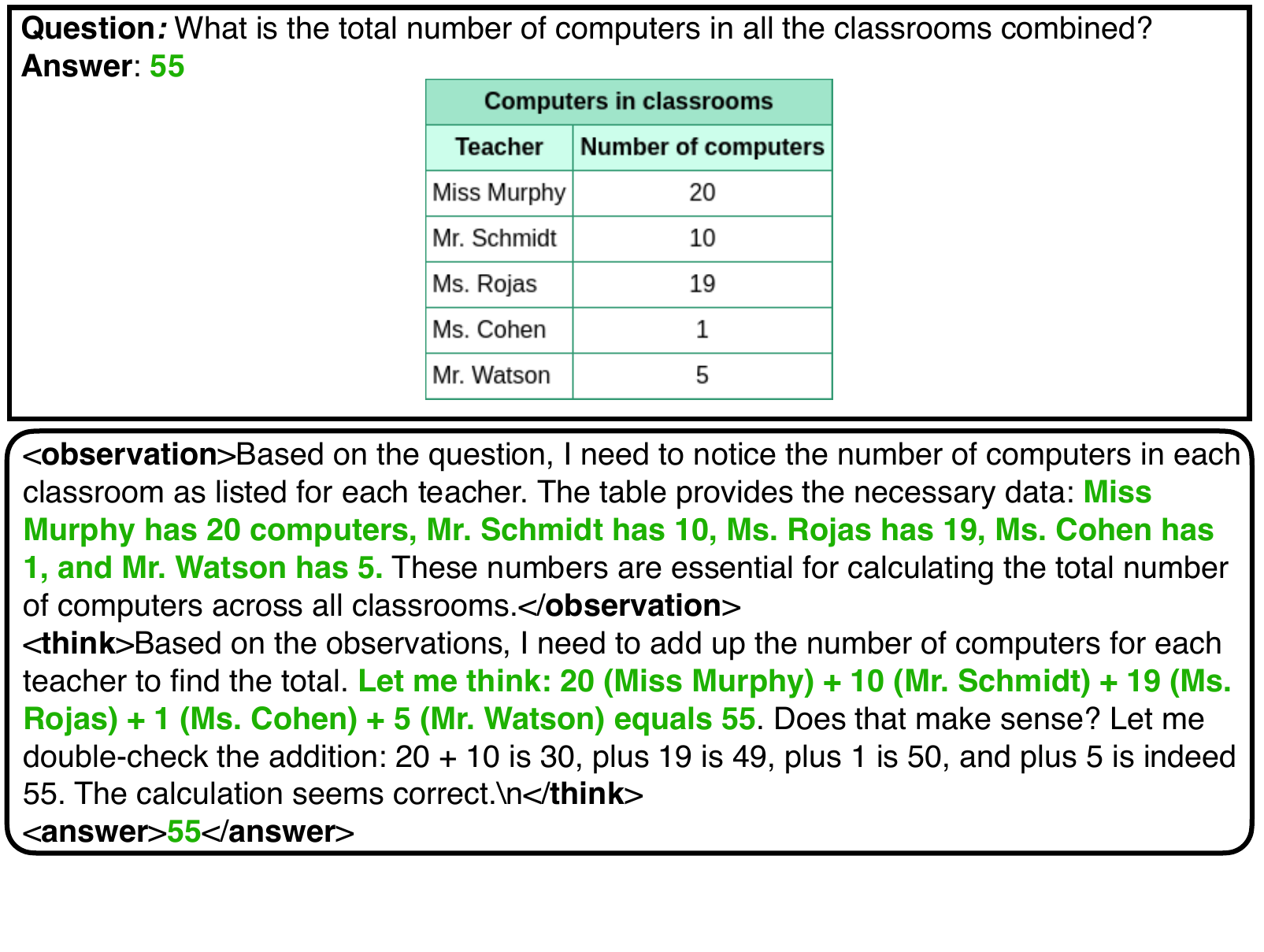}

   \caption{Annotation Example of the Image QA Sample}
   \label{fig:annotation_sample_image}
\end{figure}

We provide annotation examples in~\cref{fig:annotation_sample_video,fig:annotation_sample_image} to illustrate how our annotations explicitly separate perception from reasoning. \Cref{fig:annotation_sample_video} presents a video QA example where the perception segment successfully captures the key visual cue~(the zigzag pattern), and the reasoning segment then derives the correct answer based on this evidence. \Cref{fig:annotation_sample_image} presents an image QA example in which the perception segment accurately extracts the numerical information from the table, and the reasoning segment performs the required mathematical reasoning over these numbers, followed by validation to double-check the final answer.

%% file: sup/1_experiment_setup.tex
\section{Experiment Set up}
\label{sup:experiment_detilas}
\subsection{Implementation Details}
The whole two-stage training is conducted on 8× NVIDIA A100 GPUs.
For efficiency, we limit the video input to 16 frames at a resolution of 128 × 28 × 28 during training, where 28×28 denotes the patch size and 128 the number of patches.
For the SFT stage, we use a batch size of 8 with gradient accumulation = 2.
For the RL stage, we adopt a batch size of 56\footnote{To accelerate training, we integrate vLLM~\cite{kwon2023efficient} for sampling, assigning one GPU exclusively for sample generation and the remaining seven for model updates.}~(8 rollouts per sample). We use Claude 3.7 Sonnet for observation judgment to speed up the training process. We note that since video token processing dominates training time, the additional text-only judgment introduces little overhead: GRPO takes 16.5 hours for 1k steps, while \textsc{VideoP2R} takes 18 hours on the same hardware.

During inference, we increase the number of frames and resolution to 32 and 256 × 28 × 28, respectively, and apply the decoding configuration~(top\_p = 0.001, temperature = 0.01) consistent with the Qwen2.5-VLA official demo. During both training and inference, we adopt the same prompt~(\cref{fig:prompt_for_cot}) as in process-aware CoT generation, and use the prompt shown in \cref{fig:prompt_for_obser} for perception accuracy judgment.

\subsection{Main Table Evaluation Setup}
\label{sup:Main Table}
This section introduces the evaluation benchmarks used in \cref{tab:main_results} and the evaluation metrics.
We selected seven widely used video understanding and reasoning benchmarks to provide a comprehensive analysis of VideoP2R:
\begin{itemize}[leftmargin=* , topsep=0pt, itemsep=0pt, parsep=0pt]
\item VSI-Bench~\cite{yang2025thinking} is a video-based benchmark designed to evaluate models’ visual–spatial reasoning capability. It includes two types of questions: (1) numerical and (2) multiple-choice. Numerical questions are evaluated using Mean Relative Accuracy~(MRA), while multiple-choice questions are evaluated using Accuracy~(ACC). Following the original benchmark protocol, we report the overall performance as \textbf{the average of MRA and ACC}.

\item VideoMMMU~\cite{hu2025video} is a multi-modal, multi-disciplinary video benchmark, designed to evaluate models’ ability to acquire and apply knowledge from expert-level lecture videos. In our experiments, models are evaluated with \textbf{accuracy} over all questions.

\item MMVU~\cite{zhao2025mmvu} is an expert-level multi-disciplinary video understanding benchmark aimed at assessing models’ capability to perform domain-specific reasoning across diverse scientific and technical fields. In our experiments, models are evaluated with \textbf{accuracy} over \textbf{all multiple-choice questions}~(1858 of 3000).

\item VCR-Bench~\cite{qi2025vcr} is a benchmark crafted to assess video Chain-of-Thought reasoning. VCR-Bench selected and integrated data from multiple existing video benchmarks. In our experiments, models are evaluated with \textbf{accuracy} over all \textbf{multiple-choice questions}~(510 of 1034).

\item MVBench~\cite{li2024mvbench} is a multi-modal video understanding benchmark designed to stress test models’ temporal reasoning capabilities across diverse domains.
In our evaluation, models are assessed using \textbf{accuracy} on multiple-choice QA derived from temporally grounded tasks.

\item TempCompass~\cite{liu2024tempcompass} is a temporal reasoning benchmark designed to dissect video LLMs’ ability to perceive dynamic changes over time. It constructs paired videos that share identical static content but differ in temporal aspects (e.g., speed, direction) to prevent shortcut solutions based on static frames. In our evaluation, models are measured by \textbf{accuracy} over temporal reasoning questions under the official protocols.

\item Video-MME~\cite{fu2025video} is a comprehensive multi-modal evaluation benchmark for video-centric large language models, designed to assess their analysis capabilities across diverse video types and modalities. We evaluate using the official metrics and configuration, reporting \textbf{accuracy} over the QA pairs \textbf{without subtitles}.
\end{itemize}

For all result numbers of \textit{Open-Source Models} in \cref{tab:main_results}, we use the reported number in the original paper. For all result numbers of \textit{RFT Models}, we run the evaluation locally. We additionally include Qwen2.5-VL-72B in \cref{tab:main_results_additional} as an upper bound for our model. While \textsc{VideoP2R} still trails Qwen2.5-VL-72B on average, it significantly boosts the base model (Qwen2.5-VL-7B) and even outperforms Qwen2.5-VL-72B on MVBench, underscoring the effectiveness of our approach.

%% file: sup/2_ablation_study_on_judge_model.tex
\section{Ablation Study on Judge Model}
\label{sup:abltaion_judge}
\Cref{tab:ablation_judge} presents the results of using different judge models for perception correctness judgement. We conduct the same two-stage training process, but only change the Claude3.7 to Llama3.1~\cite{dubey2024llama} families for providing perception correctness judgment. 

Compared with the base model, all \textsc{VideoP2R} variants using different judge models achieve consistent improvements, confirming the effectiveness of perception reward supervision. To further assess judge reliability, we randomly annotate the perception correctness of 200 samples with human labels and evaluate each judge’s decision accuracy. We observe a clear upward trend in accuracy as the judge model becomes larger and more capable. Additionally, the fact that even Llama3.1-8B attains reasonable reliability on this relatively simple perception correctness judgement suggests that perception correctness can be robustly handled by current LLMs, and our pipeline is broadly applicable across a wide range of judge models. Moreover, the positive correlation between judge capability and the downstream performance of the trained model indicates that stronger judges provide more reliable perception feedback and lead to larger gains, with Claude3.7 achieving the highest agreement with human annotations and the best overall process-aware performance.

\begin{table*}[!h]
  \caption{Ablation studies of \textsc{VideoP2R} on judge models. ``Judge Acc.'' reports perception decision accuracy on 200 human-labeled samples. Best results within each group are in \textbf{bold}.}
  \centering
  \small
  \resizebox{\textwidth}{!}{%
  \begin{tabular}{l|c|rrrr|rrr|c}
    \toprule
    \multirow{2}{*}{\quad\quad Model} & \multirow{2}{*}{Judge Acc.} & \multicolumn{4}{c|}{Video Reasoning} & \multicolumn{3}{c|}{Video Understanding} & \multirow{2}{*}{Avg} \\
    \cline{3-9}
     &  & VSI. & VideoMMMU & MMVU & VCR. & MV. & TempCom. & VideoMME & \\
    \midrule
    \multicolumn{10}{l}{\textbf{• Base Models}} \\
    Qwen2.5-VL(7B)   & --  & 30.1 & 48.1 & 60.0 & 44.3 & 59.0 & 72.6 & 56.6 & 52.9 \\
    \midrule
    \multicolumn{10}{l}{\textbf{• Judge Model}} \\
    \textsc{VideoP2R}~(Llama3.1-8B)     & 82 & \textbf{39.0} & 52.2 & 64.0 & 49.2 & 64.7 & 73.8 & 59.2 & 57.4 \\
    \textsc{VideoP2R}~(Llama3.1-70B)    & 88 & 35.8 & 52.4 & 64.6 & 50.2 & 65.0 & \textbf{74.5} & \textbf{60.5} & 57.6 \\
    \textsc{VideoP2R}~(Llama3.1-405B)   & 91 & 38.2 & 54.4 & 64.5 & 49.2 & 66.5 & 75.0 & 58.4 & 58.0 \\
    \cellcolor[HTML]{E7FAFE}\textsc{VideoP2R}~(Claude3.7) &
      \cellcolor[HTML]{E7FAFE}95 &
      \cellcolor[HTML]{E7FAFE}36.8 & \cellcolor[HTML]{E7FAFE}\textbf{55.0} & \cellcolor[HTML]{E7FAFE}\textbf{65.4} &
      \cellcolor[HTML]{E7FAFE}\textbf{51.0} & \cellcolor[HTML]{E7FAFE}\textbf{68.1} &
      \cellcolor[HTML]{E7FAFE}\textbf{74.5} & \cellcolor[HTML]{E7FAFE}60.0 & \cellcolor[HTML]{E7FAFE}\textbf{58.7} \\
    \bottomrule
  \end{tabular}}
  \label{tab:ablation_judge}
\end{table*}

%% file: sup/3_observation_examination.tex
\section{Details of the Perception Examination}
\label{sup:perception examination}
\subsection{Prompt Used and Detailed Set up}
\begin{figure}[h]
  \centering
   \includegraphics[width=0.95\linewidth]{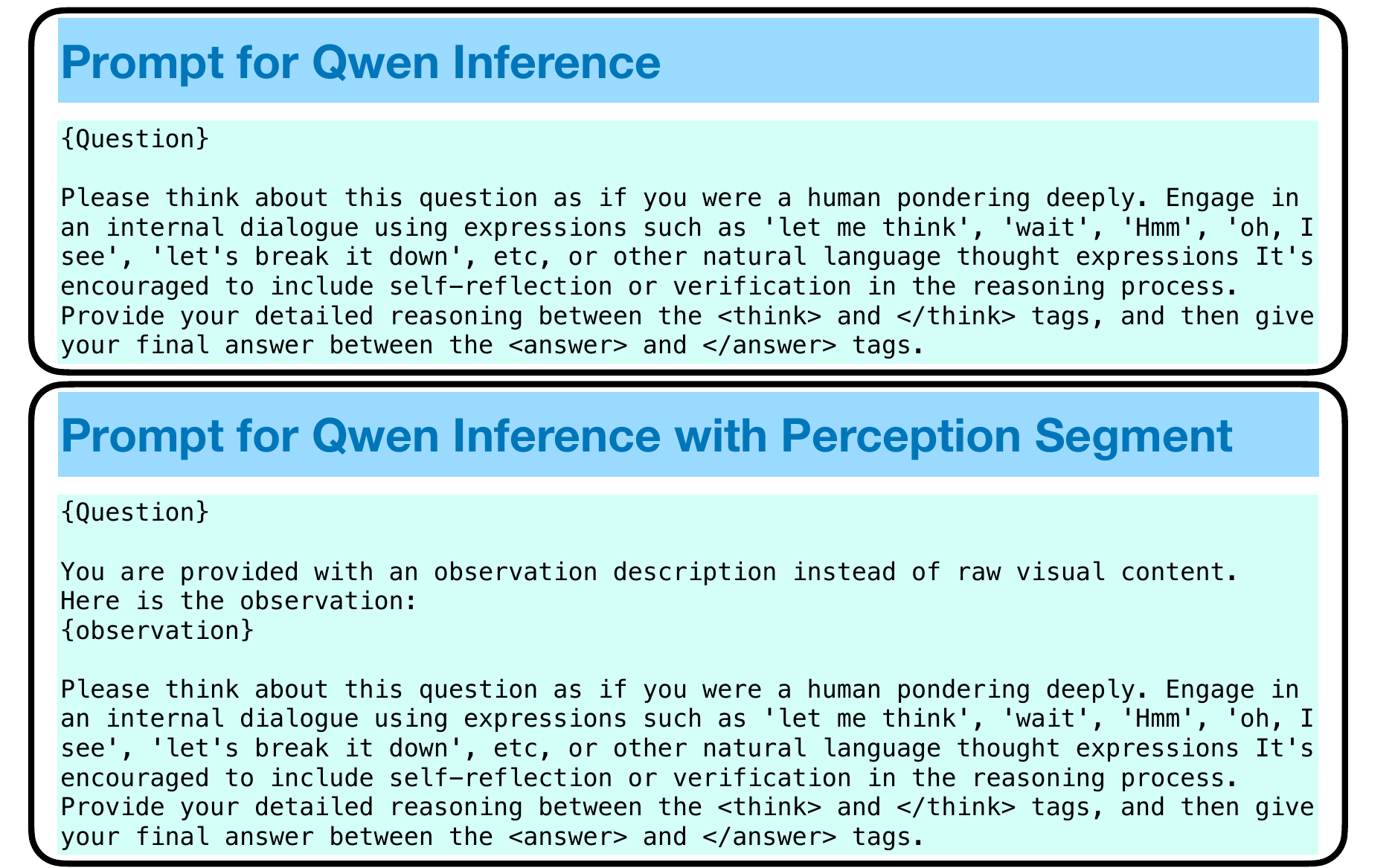}

   \caption{Prompt Template for Perception Examination Experiment.}
   \label{fig:perception_examination_template}
\end{figure}

The perception examination experiment involves three types of experiments on either text or video domains. We compare the zero-shot performance of Qwen2.5-VL-7B across different input settings and examine how perception segments influence its answers: (i) performance on text-only questions, (ii) performance with both text and video inputs, and (iii) performance when the text-only prompt is augmented with a perception segment generated by \textsc{VideoP2R} or Qwen2.5-VL-7B. 
We used the prompt ``P\textit{rompt for Qwen Inference}''~(\cref{fig:perception_examination_template} Top) for (i) and (ii). The prompt ``\textit{Prompt for Qwen Inference with Perception Segment}''~(\cref{fig:perception_examination_template} Bottom) is used for (iii). For (iii), We use the same prompt in \cref{fig:prompt_for_cot} to get the perception segment from \textsc{VideoP2R} or Qwen2.5-VL-7B first and then augment the segment within the prompt for inference. 

\subsection{Full Results}
\Cref{tab:obs_ablation} presents the full results of the perception examination experiment~(\cref{fig:observation}).

\begin{table*}[!t]
  \caption{Comparison of Qwen2.5-VL-7B with and without perception segments augmentation (video/text modalities).}
  \centering
  \small
  \resizebox{\textwidth}{!}{%
  \begin{tabular}{l|rrrr|rrr|c}
    \toprule
    \multirow{2}{*}{Model} & \multicolumn{4}{c|}{Video Reasoning} & \multicolumn{3}{c|}{Video Understanding} & \multirow{2}{*}{Avg} \\
    \cline{2-8}
     & VSI. & VideoMMMU & MMVU & VCR. & MV. & TempCom. & VideoMME & \\
    \midrule
    \multicolumn{9}{l}{\textbf{• Video + Text}} \\
    Qwen2.5-VL(7B)                  & 30.1 & 48.1 & 60.0 & 44.3 & 59.0 & 72.6 & 56.6 & 52.9 \\
    Qwen2.5-VL(7B) + Obs.           & 29.7 & 54.1 & 61.5 & 49.4 & 65.4 & 73.4 & 58.8 & 56.0 \\
    Qwen2.5-VL(7B) + Obs.(Qwen)     & 30.4 & 53.1 & 61.5 & 48.6 & 61.6 & 72.8 & 56.4 & 54.9 \\
    \midrule
    \multicolumn{9}{l}{\textbf{• Text }} \\
    Qwen2.5-VL(7B)                  & 22.1 & 34.4 & 45.8 & 33.5 & 30.9 & 46.2 & 33.1 & 35.1 \\
    Qwen2.5-VL(7B) + Obs.           & 33.8 & 51.5 & 61.0 & 48.4 & 63.6 & 72.9 & 57.5 & 55.5 \\
    Qwen2.5-VL(7B) + Obs.(Qwen)     & 28.8 & 45.8 & 60.0 & 41.5 & 49.1 & 67.3 & 48.0 & 48.6 \\
    \midrule
    \cellcolor[HTML]{E7FAFE}\textsc{VideoP2R} & 
      \cellcolor[HTML]{E7FAFE}\textbf{36.8} & 
      \cellcolor[HTML]{E7FAFE}\textbf{55.0} & 
      \cellcolor[HTML]{E7FAFE}\textbf{65.4} & 
      \cellcolor[HTML]{E7FAFE}\textbf{51.0} & 
      \cellcolor[HTML]{E7FAFE}\textbf{68.1} & 
      \cellcolor[HTML]{E7FAFE}\textbf{74.5} & 
      \cellcolor[HTML]{E7FAFE}\textbf{60.0} & 
      \cellcolor[HTML]{E7FAFE}\textbf{58.7} \\
    \bottomrule
  \end{tabular}}
  \label{tab:obs_ablation}
\end{table*}

\begin{figure}[h]
  \centering
   \includegraphics[width=0.95\linewidth]{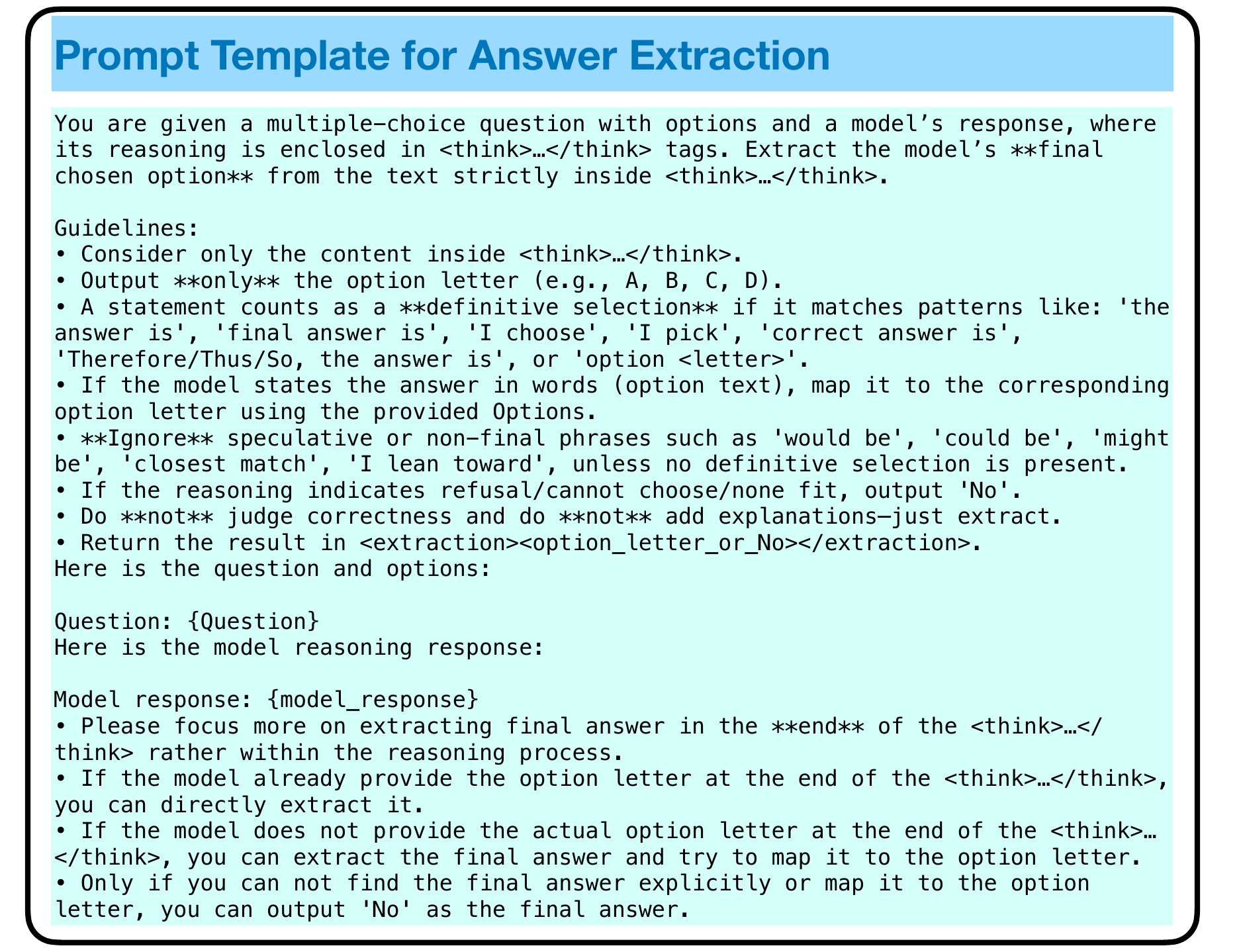}

   \caption{Prompt Template for Answer Extraction.}
   \label{fig:prompt_for_answer_extraction}
\end{figure}

\subsection{Examples of Qwen Inference Output}

We present examples of Qwen’s outputs under different configurations in our perception examination experiment in \cref{fig:perception_examples}. When given only the text question~(Top Left), Qwen fails to perform meaningful reasoning due to the absence of video information and resorts to guessing from the answer choices. When conditioned on the text question plus Qwen’s own perception segment~(Top Right), it mentions some relevant visual cues~(e.g., the person placing books into the backpack) but omits critical details such as the exact number of books, resulting in unreliable reasoning. Even with access to both the text question and the video input~(Bottom Left), Qwen still produces inaccurate perception, confusing the top pocket with the main compartment; this misperception propagates into the reasoning process and yields an incorrect answer. In contrast, the perception segment generated by \textsc{VideoP2R} is clear and sufficient~(Bottom Right), explicitly capturing both the number of books and their correct placement. This improvement can be attributed to PA-GRPO, which guides the perception process to target information directly relevant to answering the question, thereby enabling Qwen to arrive at the correct answer even without direct access to the video.

%% file: sup/4_rl_training_trend.tex
\section{RL Training Dynamics}
\label{sup:rl training trend}

\begin{figure}[h]
  \centering
   \includegraphics[width=0.9\linewidth]{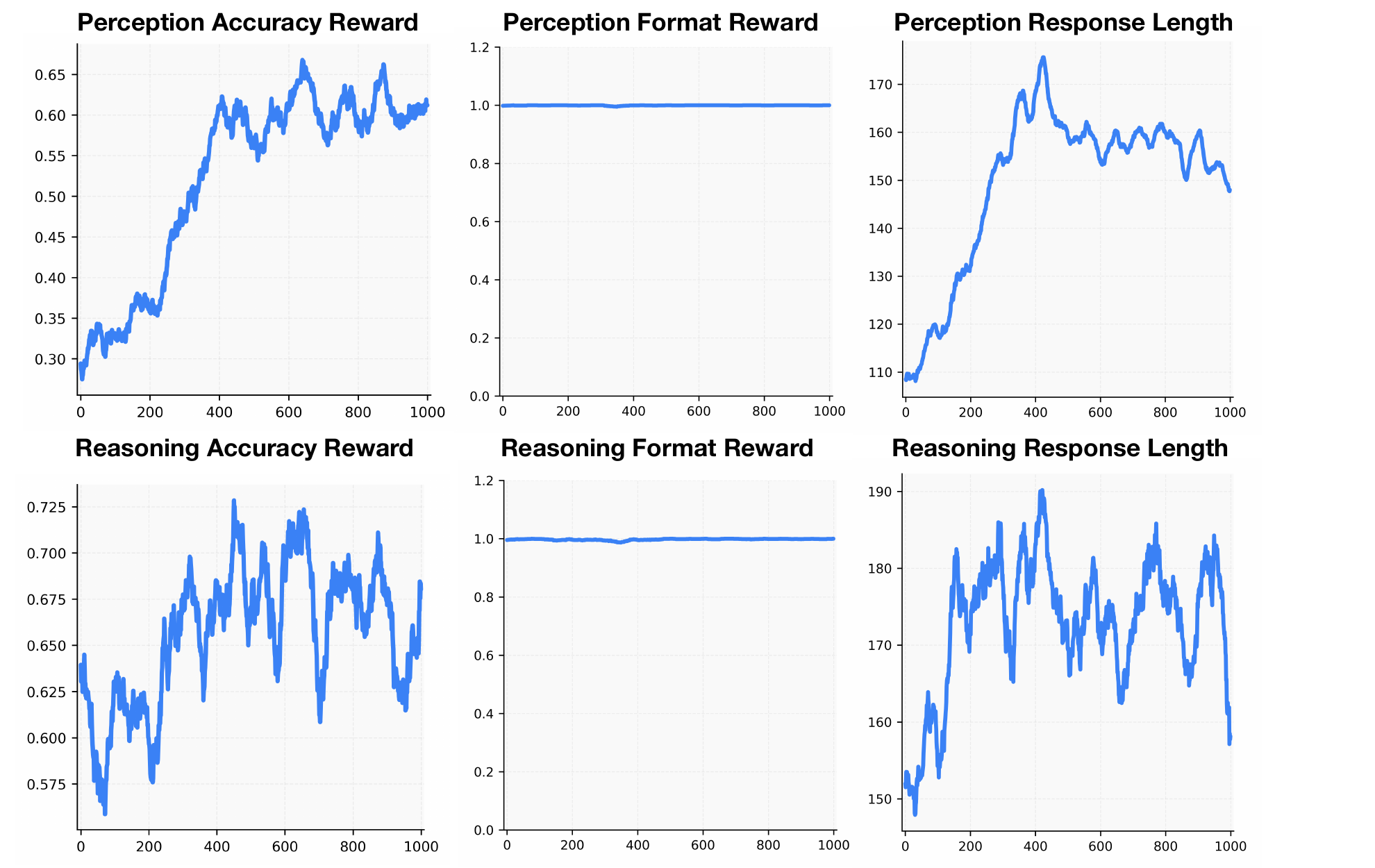}

   \caption{RL training Dynamics of \textsc{VideoP2R}}
   \label{fig:reward_trend}
\end{figure}

We provide the full RL training dynamics of \textsc{VideoP2R} in \cref{fig:reward_trend} to comprehensively illustrate our RL stage. Both the perception accuracy reward and the reasoning accuracy reward exhibit an overall increasing trend, indicating that the model progressively improves its ability to produce correct perception and reasoning traces. The perception format reward and reasoning format reward remain close to 1 throughout training, showing that the model consistently adheres to the process-aware inference template and maintains stable format compliance. Since the length reward is conditioned on both accuracy and format rewards, we instead visualize the lengths of the perception and reasoning segments during RL. We observe an initial increase followed by a decrease in both segments, indicating that the model adaptively adjusts its outputs and eventually converges to concise yet sufficiently informative perception and reasoning traces. 

%% file: sup/5_alignment_check.tex
\section{Think-Answer Mismatch Analysis}
\label{sup:think-answer}
\subsection{Pilot Experiment}

We conduct a pilot experiment with Claude 3.7 to assess the reliability of answer extraction using the prompt in \cref{fig:prompt_for_answer_extraction}. We first sample 400 responses to multiple-choice questions and ask human annotators to extract the model’s answers from the $\langle \texttt{think} \rangle$ segments. The annotators achieve 95\% agreement, with most disagreements arising from cases where the reasoning is unclear and the model appears to guess the answer. Using the same setting, Claude 3.7 reaches 96.5\% accuracy, confirming the reliability of this assessment.

\subsection{Think-Answer Mismatch Example}

\begin{figure}[h]
  \centering
   \includegraphics[width=0.95\linewidth]{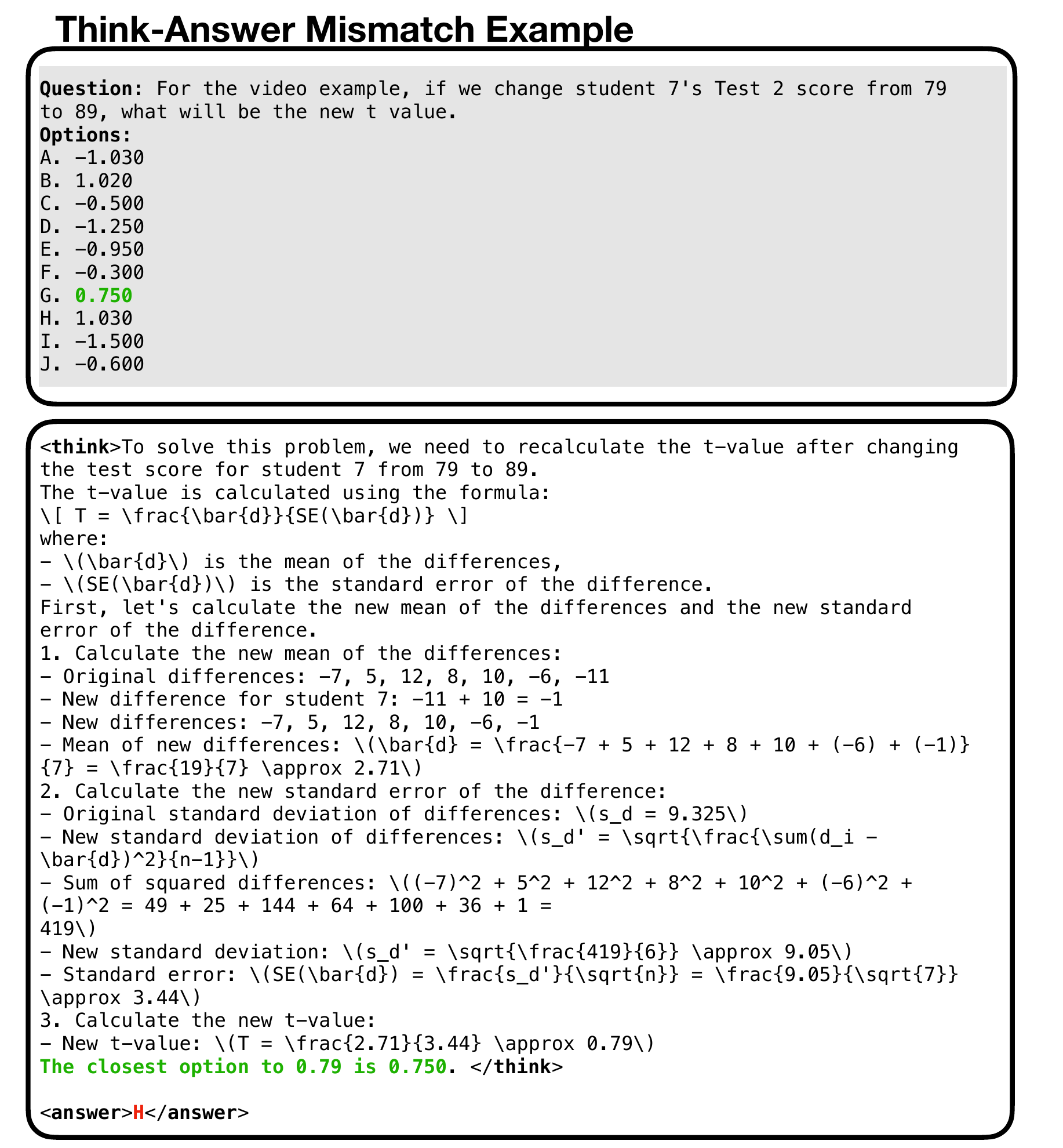}

   \caption{Example of Think-Answer Mismatch.}
   \label{fig:example_think_answer_mismatch}
\end{figure}

\Cref{fig:example_think_answer_mismatch} presents an example of Think–Answer Mismatch, where the model conducts correct reasoning in $\langle \texttt{think} \rangle$ but produces an incorrect final answer in $\langle \texttt{answer} \rangle$. This mismatch highlights how relying solely on final-answer rewards can reinforce unfaithful or inconsistent behavior, underscoring the necessity of process-aware rewards in PA-GRPO.

\begin{table*}[!t]
    \caption{Think–Answer Mismatch rates across models and benchmarks~(lower is better).}
  \centering
  \small
  \resizebox{\textwidth}{!}{%
  \begin{tabular}{l|rrrr|rrr|c}
    \toprule
    \multirow{2}{*}{Model} & \multicolumn{4}{c|}{Video Reasoning} & \multicolumn{3}{c|}{Video Understanding} & \multirow{2}{*}{Avg} \\
    \cline{2-8}
     & VSI. & VideoMMMU & MMVU & VCR. & MV. & TempCom. & VideoMME & \\
    \midrule
    Qwen2.5-VL(7B)   & 4.9 & 12.2 & 3.5 & 6.3 & 7.8 & 4.6 & 5.1 & 6.3 \\
    Video-R1 (SFT)   & 2.0 & 2.7  & 3.2 & 1.0 & 0.6 & 0.5 & 0.8 & 1.6 \\
    VideoRFT (SFT)   & 5.4 & 4.2  & 9.1 & 5.0 & 7.3 & 1.6 & 3.6 & 5.2 \\
    VideoP2R (SFT)   & 0.9 & 4.2  & 0.9 & 1.0 & 1.4 & 0.9 & 1.2 & 1.5 \\
    Video-R1      & 25.5 & 19.0 & 18.8 & 26.7 & 21.7 & 14.5 & 24.3 & 21.5 \\
    VideoChat-R1 & 13.6 & 11.9 & 11.9 & 13.3 & 12.4 & 7.3  & 11.3 & 11.7 \\
    VersaVid-R1*   & --   & 21.5 & 11.1 & 18.6 & 16.7 & 9.5  & 15.6 & 15.5 \\
    VideoRFT     & 23.5 & 22.9 & 12.9 & 18.5 & 15.1 & 10.9 & 14.4 & 16.9 \\
    \cellcolor[HTML]{E7FAFE}\textsc{VideoP2R} &
      \cellcolor[HTML]{E7FAFE}6.8 &
      \cellcolor[HTML]{E7FAFE}9.6 &
      \cellcolor[HTML]{E7FAFE}7.9 &
      \cellcolor[HTML]{E7FAFE}7.6 &
      \cellcolor[HTML]{E7FAFE}7.4 &
      \cellcolor[HTML]{E7FAFE}6.4 &
      \cellcolor[HTML]{E7FAFE}7.1 &
      \cellcolor[HTML]{E7FAFE}7.5 \\
    \bottomrule
  \end{tabular}}
  \label{tab:mismatch}
\end{table*}

\subsection{Think-Answer Mismatch Results}
\Cref{tab:mismatch}\footnote{*VersaVid-R1 has too few available traces on VSI-Bench for meaningful statistics.} reports Think–Answer Mismatch rates across benchmarks. All results are computed on the multiple-choice subsets of each benchmark.

%% file: sup/6_more_cases.tex
\section{More Qualitative Results of \textsc{VideoP2R}}
\label{sup:more_qualitative results}

\subsection{Success Case}
\begin{figure*}[h]
  \centering
   \includegraphics[width=0.9\linewidth]{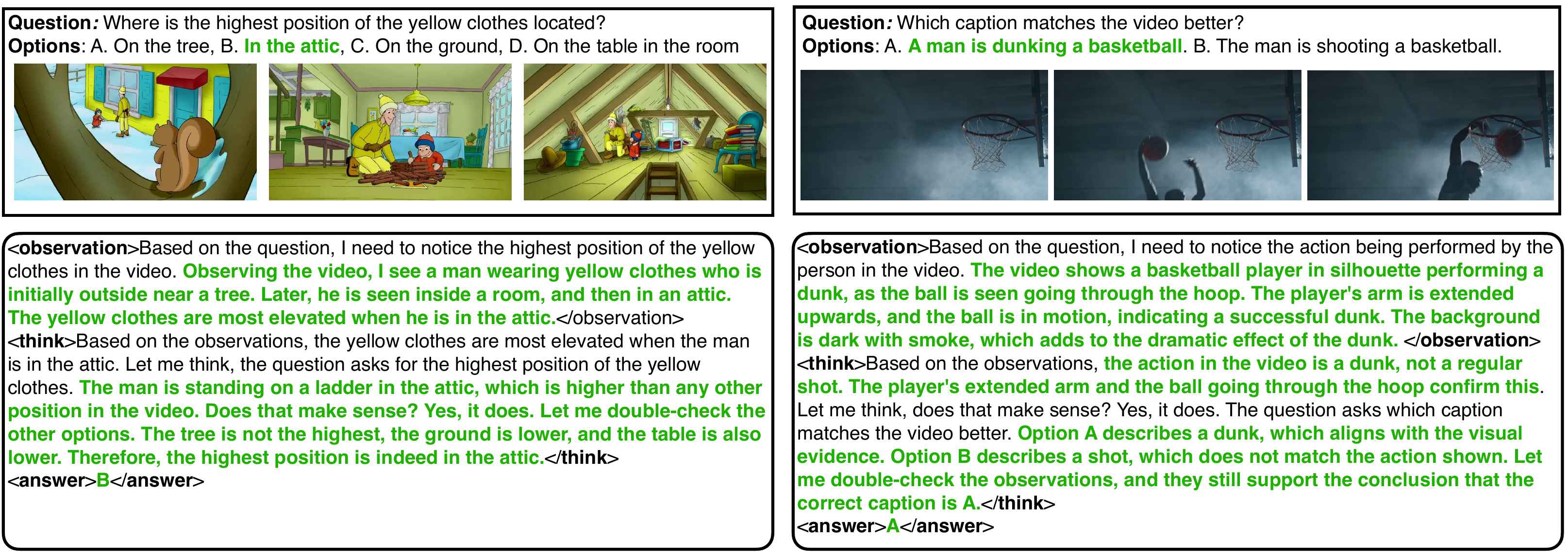}

   \caption{Success Cases of \textsc{VideoP2R}}
   \label{fig:success_case_more}
\end{figure*}

We provide two additional success cases of \textsc{VideoP2R} in \cref{fig:success_case_more}. In the left example, \textsc{VideoP2R} effectively tracks key visual information throughout the video: in the $\langle\texttt{observation}\rangle$ segment, it identifies the three positions where the yellow clothing is presented, supporting subsequent reasoning. In the right example, \textsc{VideoP2R} accurately captures relevant visual cues, including the person’s gestures and the background context, and leverages them to produce the correct final answer.

\subsection{Failure Case}

\begin{figure}[h]
  \centering
   \includegraphics[width=0.9\linewidth]{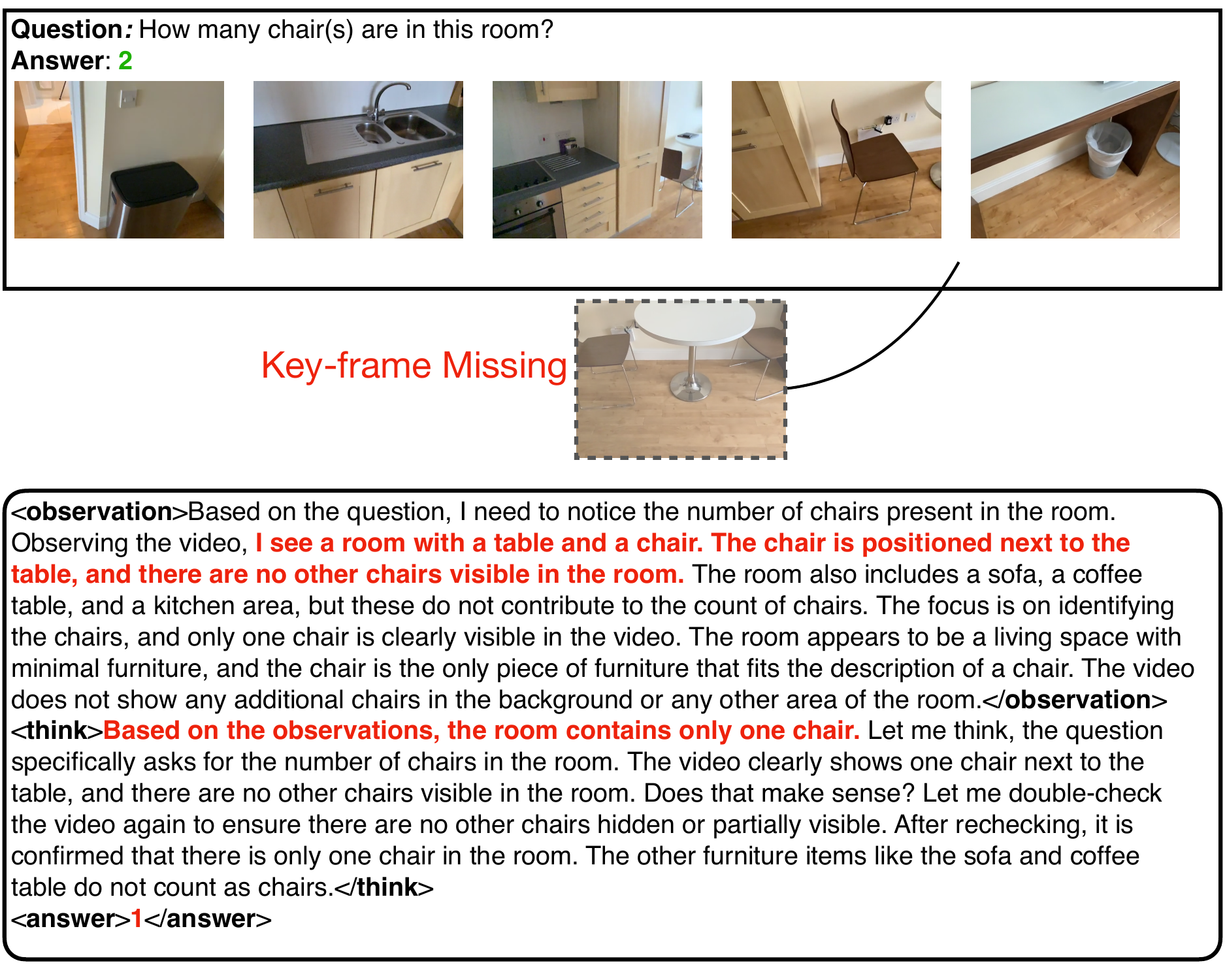}

   \caption{Failure Cases of Key-frame Missing in \textsc{VideoP2R}}
   \label{fig:failure_case_more_1}
\end{figure}

\begin{figure}[h]
  \centering
    \includegraphics[width=0.9\linewidth]{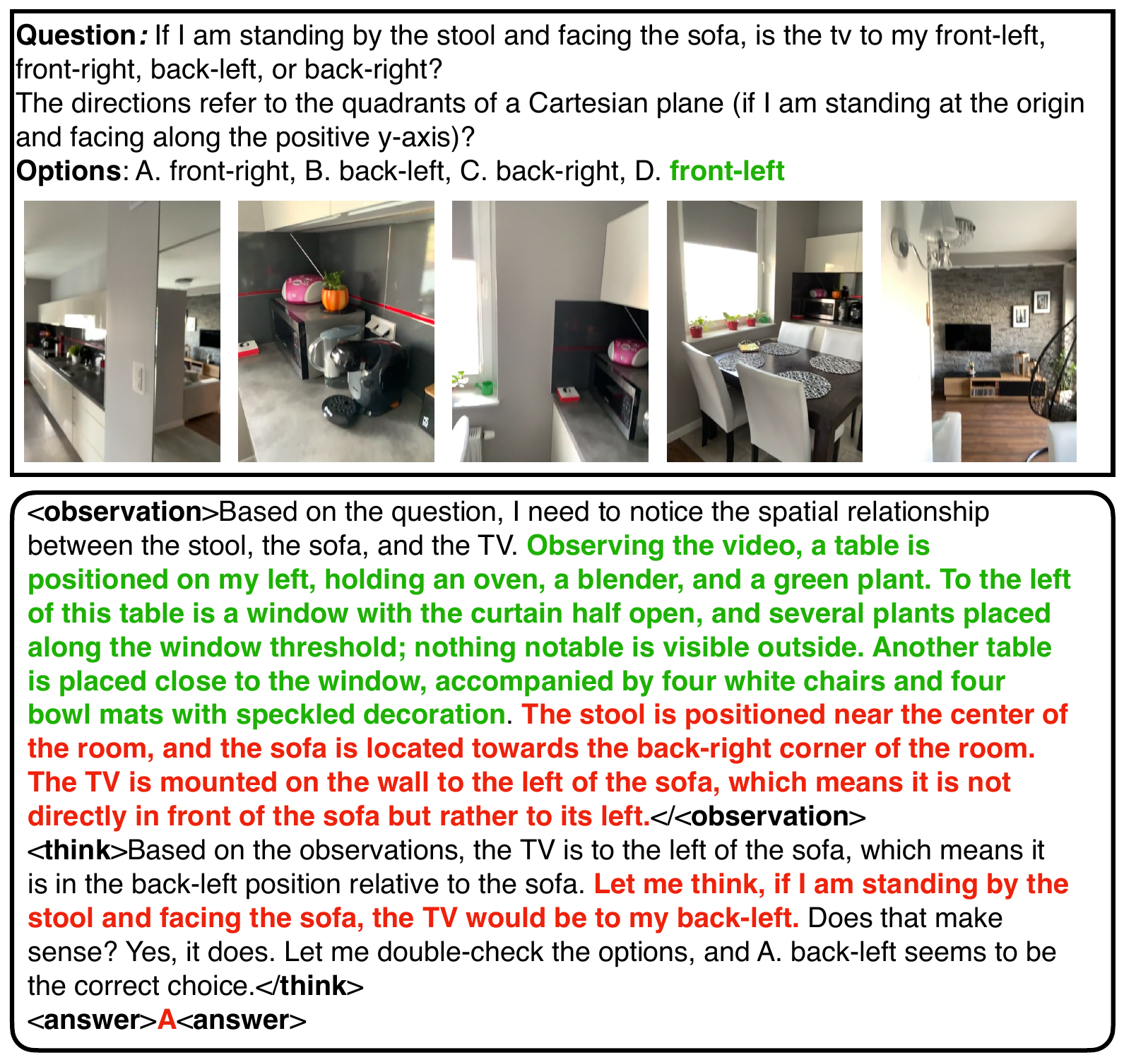}

   \caption{Failure Cases of Overly detailed visual configuration in \textsc{VideoP2R}}
   \label{fig:failure_case_more_2}
\end{figure}

We identify two representative types of failure cases for \textsc{VideoP2R}, illustrated in \cref{fig:failure_case_more_1} and \cref{fig:failure_case_more_2}. \textbf{(1) Key-frame missing.} During inference, we adopt uniform frame sampling, which may omit question-critical key frames containing essential visual evidence~\cite{chasmai2025moment}. As shown in \cref{fig:failure_case_more_1}, the question asks for the number of chairs in the room. However, the sampled frames only include the view where one chair appears on the left side of the table, while the key frame showing another chair on the right side is missed. In this case, the question becomes unsolvable given the incomplete observations. Increasing the number of sampled frames can mitigate this issue, and an adaptive sampling strategy~\cite{han2024self} can further reduce the risk of missing question-critical evidence, which we leave for future work. \textbf{(2) Overly detailed visual configuration.} The second failure type arises when questions require tracking an excessive number of fine-grained visual details, which is particularly common in VSI-Bench~\cite{yang2025thinking}.  Questions in VSI-Bench demand precise modeling of object layouts and relative positions across multiple regions, often exceeding the length targets~([128,320]) used during \textsc{VideoP2R}'s training. When the required descriptions fall outside this familiar length regime, the model tends to compress or drop critical perceptual details, leading to incomplete observations and subsequent reasoning errors. As shown in \cref{fig:failure_case_more_2}, the question targets the relative positions among the stool, sofa, and TV, while accurately specifying this configuration also requires the locations of surrounding reference objects, such as the table near the stool and the table next to the window. \textsc{VideoP2R} initially exhibits high-quality perception (e.g., correctly identifying the tables and window) but gradually introduces errors for later elements (e.g., the stool and sofa), which ultimately leads to failure in subsequent reasoning. We propose to have a more dynamic length reward system~\cite{wan2025srpo} in future work.

\subsection{From Base Model to \textbf{\textsc{VideoP2R}}: A Stepwise Capability Evolution}

\begin{figure*}[h]
  \centering
   \includegraphics[width=0.9\linewidth]{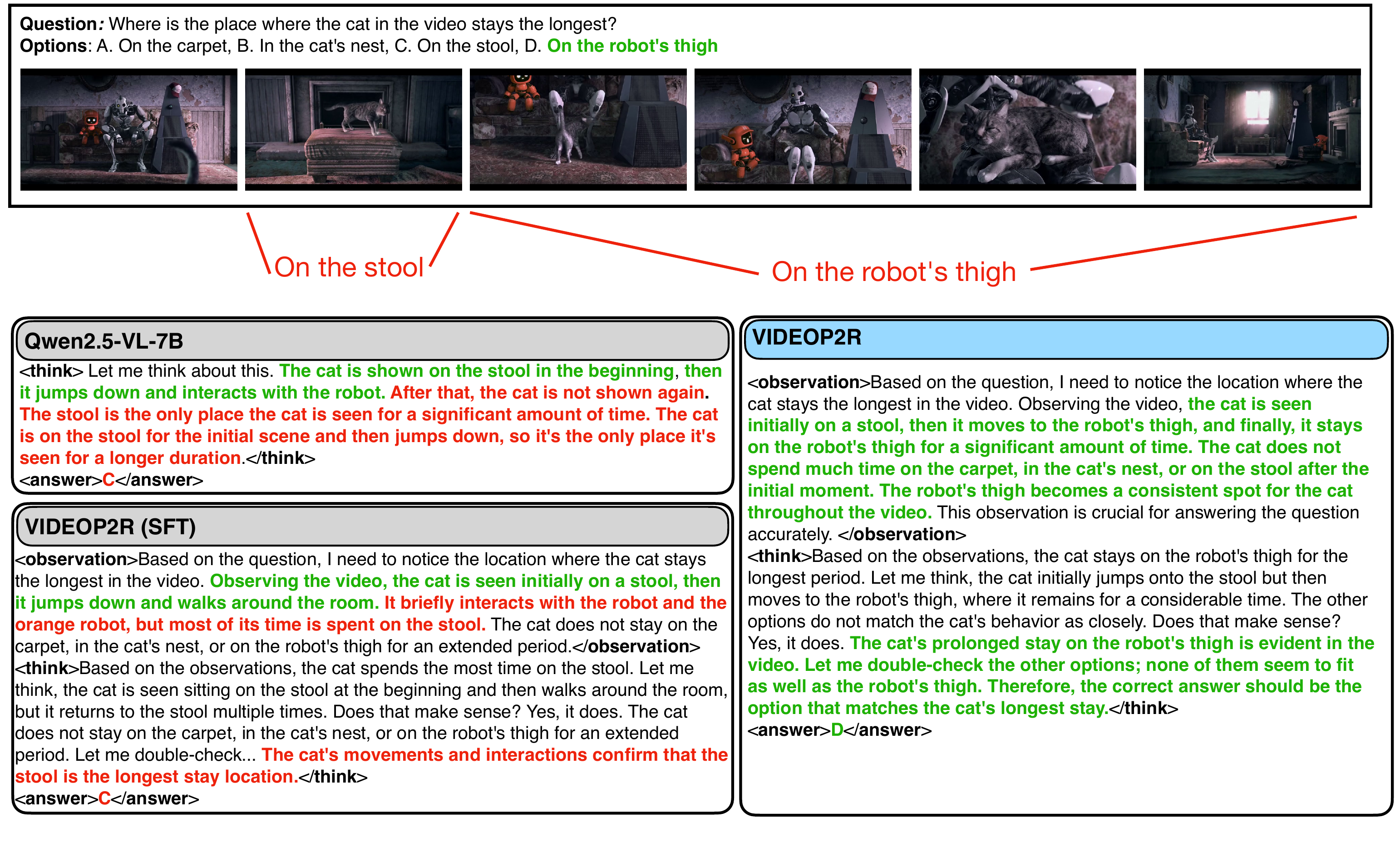}

   \caption{From base Qwen2.5-VL-7B to \textsc{VideoP2R}-SFT and \textsc{VideoP2R}: a representative example illustrating the stepwise improvement in model's perception and reasoning.}
   \label{fig:different_stage}
\end{figure*}

We present a representative example in \cref{fig:different_stage} to illustrate how the model’s perception and reasoning capabilities evolve from the base Qwen2.5-VL-7B to \textsc{VideoP2R-SFT} and finally to the RL-optimized \textsc{VideoP2R}. The question asks where the cat stays for the longest time. In the video the cat briefly starts on the stool and then spends the remaining time on the robot’s thigh. The base Qwen2.5-VL-7B only captures the early details where the cat is on the stool and misses the later details on the robot’s thigh, leading to an incorrect answer. \textsc{VideoP2R-SFT} attends to both the stool and the robot’s thigh but misjudges the relative durations. In contrast, \textsc{VideoP2R} produces a comprehensive and faithful perception trace that correctly tracks the cat’s locations and time spent, even explicitly ruling out distractors such as the carpet and nest, thereby enabling reliable downstream reasoning. This example demonstrates that each stage~(SFT, RL) of the \textsc{VideoP2R} framework is both effective and necessary, enabling stable, long-term improvement in process-aware perception and reasoning.

\subsection{Impact of Task Difficulty}
We analyze \textsc{VideoP2R}'s adaptability across difficulty levels using the fine-grained categorizations of our benchmarks~\cref{sup:Main Table}. VideoMMMU~\cite{hu2025video} spans three reasoning difficulties from basic knowledge identification (\textit{Identify}) to interpretation (\textit{Interpret}) and adaptation (\textit{Adapt}), while Video-MME~\cite{fu2025video} categorizes perception difficulty by video duration (\textit{Short}, \textit{Medium}, \textit{Long}). As shown in~\cref{tab:difficulty}, \textsc{VideoP2R} obtains its largest gains on the more challenging splits: +3.66\% on \textit{Interpret} (VideoMMMU) and +3.24\% on \textit{Medium} videos (Video-MME), while the smallest gains occur on the easiest splits (+2.31\% on \textit{Identify}, +1.66\% on \textit{Short}). These results indicate that our method is most beneficial for challenging tasks, while providing moderate gains for simpler tasks that already fall within the model's intrinsic capability boundary.

\begin{table*}[!t]
  \caption{Performance across difficulty levels on Video-MME (perception, by video duration) and VideoMMMU (reasoning). Best result of each column is in \textbf{bold} (all numbers in \%).}
  \vspace{-2mm}
  \centering
  \small
  \setlength{\tabcolsep}{5pt}
  \renewcommand{\arraystretch}{0.98}
  \begin{tabular}{l|rrr|rrr}
    \toprule
    \multirow{2}{*}{Model} & \multicolumn{3}{c|}{Video-MME (Perception)} & \multicolumn{3}{c}{VideoMMMU (Reasoning)} \\
    \cmidrule(lr){2-4}\cmidrule(lr){5-7}
    & Short & Medium & Long & Identify & Interpret & Adapt \\
    \midrule
    Qwen2.5-VL-7B & 68.49 & 55.92 & 51.06 & 68.67 & 48.67 & 42.00 \\
    \rowcolor[HTML]{E7FAFE}
    \textsc{VideoP2R} & \textbf{70.80} & \textbf{59.16} & \textbf{53.65} & \textbf{70.33} & \textbf{52.33} & \textbf{44.33} \\
    \midrule
    $\Delta$ & +2.31 & +3.24 & +2.59 & +1.66 & +3.66 & +2.33 \\
    \bottomrule
  \end{tabular}
  \vspace{-2mm}
  \label{tab:difficulty}
\end{table*}

\begin{figure*}[h]
  \centering
   \includegraphics[width=0.95\linewidth]{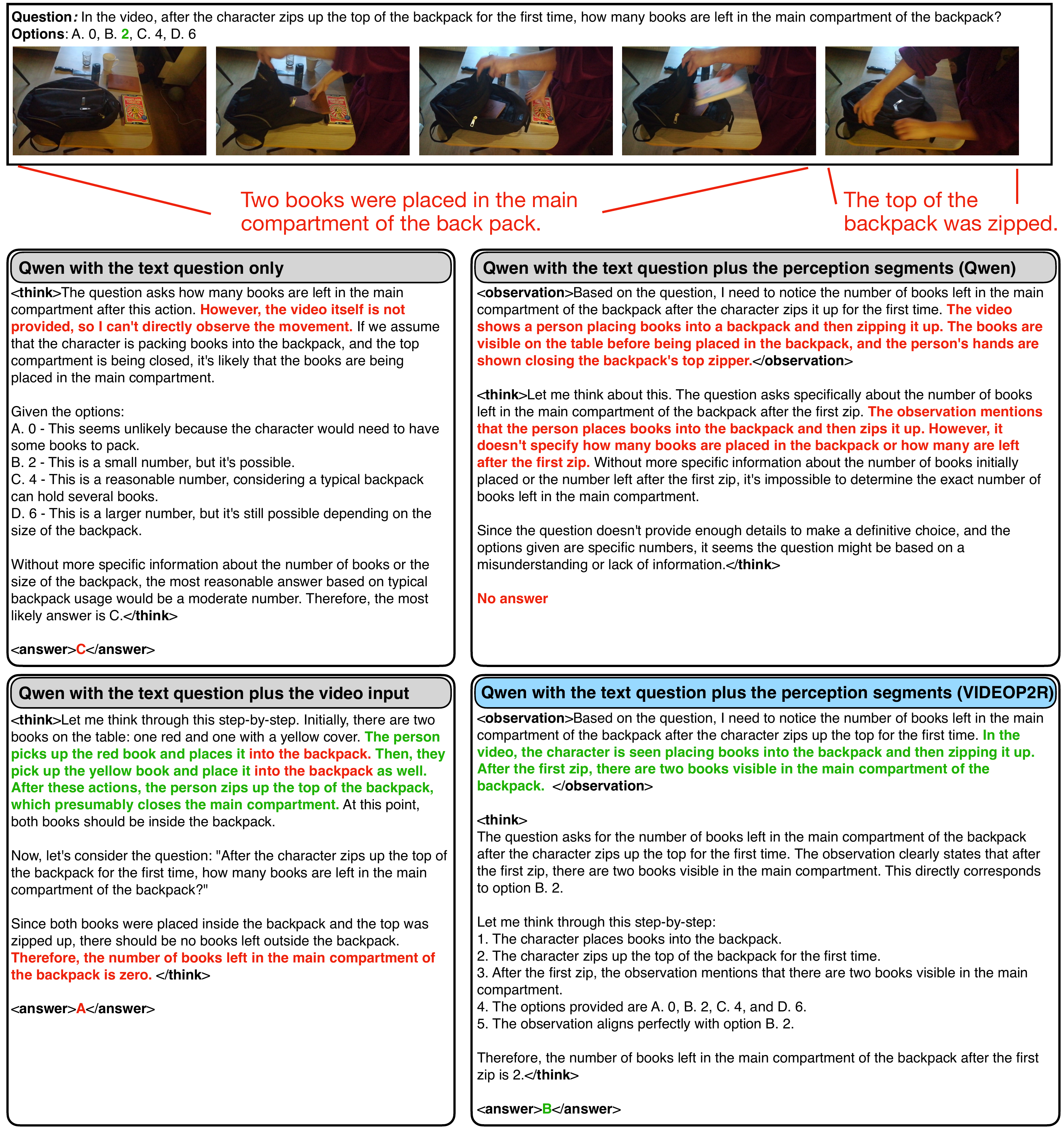}

   \caption{Examples of Perception Examination: Top Left: Qwen with the text question only; Top Right: Qwen with the text question plus the perception segments from Qwen; Bottom Left: Qwen with the text question plus the video input; Bottom Right: Qwen with the text question plus the perception segments from \textsc{VideoP2R}. \textcolor{ForestGreen}{Green} text denotes correct visual information or reasoning traces, while \textcolor{red}{red} text denotes incorrect or insufficient visual information or reasoning traces.}
   \label{fig:perception_examples}
\end{figure*}

\begin{table*}[!t]
  \caption{Performance comparison on video reasoning and understanding benchmarks. Best/second-best result of each column is in \textbf{bold}/\underline{underline}. Missing entries indicate unreported results (all numbers unit in \%).}
  \vspace{-2mm} %
  \centering
  \small
  \setlength{\tabcolsep}{6pt}
  \renewcommand{\arraystretch}{0.98}
  \begin{tabular}{l|rrrr|rrr|c}
    \toprule
    \multirow{2}{*}{Model} & \multicolumn{4}{c|}{Video Reasoning} & \multicolumn{3}{c|}{Video Understanding} & \multirow{2}{*}{Avg} \\
    \cmidrule(lr){2-5}\cmidrule(lr){6-8}
    & VSI. & VideoMMMU & MMVU & VCR. & MV. & TempCom. & VideoMME & \\
    \midrule
    Qwen2.5-VL-72B       & 37.2 & 67.0 & 73.4 & 54.9 & 61.7 & 74.9 & 64.5 & 61.9 \\
    \midrule
    \rowcolor[HTML]{F8F9FB}
    \multicolumn{9}{l}{\textbf{Open-Source 7B Models}}\\[-1pt]
    LLaVA-OneVision-7B  & 32.4 & 33.8 & 49.2 & -- & 56.7 & --  & 58.2 & -- \\
    LongVA-7B           & 29.2 & 23.9 & --   & -- & --   & 56.9 & 52.6 & -- \\
    Video-UTR-7B        & --   & --   & --   & -- & 58.8 & 59.7 & 52.6 & -- \\
    VideoLLaMA2-7B      & --   & --   & 44.8 & -- & 54.6 & --   & 47.9 & -- \\
    Qwen2.5-VL-3B       & 26.9 & 34.3 & 51.5 & 35.1 & 47.0 & 23.5 & 50.7 & 38.4 \\
    Qwen2.5-VL-7B       & 30.1 & 48.1 & 60.0 & 44.3 & 59.0 & 72.6 & 56.6 & 52.9 \\
    \midrule
    \rowcolor[HTML]{F8F9FB}
    \multicolumn{9}{l}{\textbf{RFT on Qwen2.5-VL-7B}}\\[-1pt]
    Video\text{-}R1        & \underline{35.8} & 52.3 & 63.8 & 49.0 & 63.9 & 73.2 & 59.3 & \textit{56.8} \\
    VideoChat\text{-}R1    & 33.9 & \underline{54.0} & 63.0 & 49.0 & \underline{67.9} & 72.5 & 57.7 & \textit{56.9} \\
    Time\text{-}R1         & 29.0 & 51.0 & 62.9 & 49.6 & 63.1 & 73.7 & 59.3 & \textit{55.5} \\
    VersaVid\text{-}R1     & 33.7 & 51.9 & 64.3 & \underline{49.8} & 62.9 & \underline{74.0} & 58.8 & \textit{56.5} \\
    VideoRFT               & \textbf{36.8} & 51.1 & \textbf{68.5} & 49.6 & 62.1 & 73.7 & \underline{59.8} & \underline{\textit{57.4}} \\
    \midrule
        \textsc{VideoP2R~(3B)} (Ours)      & 38.7 & 45.3 & 58.7 & 45.3 & 63.2 & 66.4 & 55.9 & 53.3 \\
    \rowcolor[HTML]{E7FAFE}
    \textsc{VideoP2R} (Ours)      & \textbf{36.8} & \textbf{55.0} & \underline{65.4} & \textbf{51.0} & \textbf{68.1} & \textbf{74.5} & \textbf{60.0} & \textbf{\textit{58.7}} \\
    \bottomrule
  \end{tabular}
  \vspace{-2mm} %
  \label{tab:main_results_additional}
  \vspace{-1mm} %
\end{table*}

%% file: sup/7_dataset_ablation.tex
\section{Impact of Model Size and Dataset Composition}
\label{sup:dataset_contribution}

We report the full results in~\cref{tab:main_results_additional}, including \textsc{VideoP2R}'s adaptation to a smaller model. \textsc{VideoP2R} also scales effectively to smaller architectures: applying our pipeline to Qwen2.5-VL-3B improves average accuracy from 38.4\% to 53.3\%, surpassing even the Qwen2.5-VL-7B baseline (52.9\%).

18:44To demonstrate that \textsc{VideoP2R}-CoT-162K is beneficial and generalizes to other RFT pipelines, we compare it against Video-R1-CoT-165K, the CoT dataset used for SFT in Video-R1~\cite{feng2025video}. Specifically, in the SFT stage, we train Qwen2.5-VL-7B on each dataset separately, and in the RL stage, we apply Video-R1's temporal GRPO~(T-GRPO) on top of both SFT checkpoints under identical settings. As shown in~\cref{tab:dataset_contribution}, training on \textsc{VideoP2R}-CoT-162K yields 55.6\% average accuracy in the SFT stage, outperforming Video-R1-CoT-165K (53.9\%) by 1.7\%. After applying T-GRPO, our dataset further improves to 57.0\%, still surpassing the Video-R1 counterpart (56.8\%). These results suggest that \textsc{VideoP2R}-CoT-162K not only provides a stronger foundation in the SFT stage, but also generalizes well to other RFT pipelines.

\begin{table*}[!t]
    \caption{Comparison between \textsc{VideoP2R}-CoT-162K and Video-R1-CoT-165K (all numbers in \%). Best result of each column is in \textbf{bold}.}
  \vspace{-2mm}
  \centering
  \small
  \setlength{\tabcolsep}{4pt}
  \renewcommand{\arraystretch}{0.98}
  \begin{tabular}{l|rrrr|rrr|c}
    \toprule
    \multirow{2}{*}{Method} & \multicolumn{4}{c|}{Video Reasoning} & \multicolumn{3}{c|}{Video Understanding} & \multirow{2}{*}{Avg} \\
    \cmidrule(lr){2-5}\cmidrule(lr){6-8}
    & VSI. & VideoMMMU & MMVU & VCR. & MV. & TempCom. & VideoMME & \\
    \midrule
    \rowcolor[HTML]{F8F9FB}
    \multicolumn{9}{l}{\textbf{SFT Stage}}\\[-1pt]
    SFT (Video-R1-CoT-165K)            & 33.3 & 49.4 & \textbf{63.5} & 45.5 & 60.5 & 69.9 & 55.4 & 53.9 \\
    \rowcolor[HTML]{E7FAFE}
    SFT (\textsc{VideoP2R}-CoT-162K)           & \textbf{35.2} & \textbf{53.7} & 61.6 & \textbf{46.9} & \textbf{62.3} & \textbf{72.4} & \textbf{57.2} & \textbf{55.6} \\
    \midrule
    \rowcolor[HTML]{F8F9FB}
    \multicolumn{9}{l}{\textbf{RL Stage}}\\[-1pt]
    SFT+T\text{-}GRPO (Video-R1-CoT-165K)  & 35.8 & \textbf{52.3} & \textbf{63.8} & \textbf{49.0} & 63.9 & 73.2 & \textbf{59.3} & 56.8 \\
    \rowcolor[HTML]{E7FAFE}
    SFT+T\text{-}GRPO (\textsc{VideoP2R}-CoT-162K) & \textbf{39.9} & 50.0 & 62.8 & 48.9 & \textbf{64.5} & \textbf{73.5} & 58.2 & \textbf{57.0} \\
    \bottomrule
  \end{tabular}
  \vspace{-2mm}
  \label{tab:dataset_contribution}
\end{table*}